# Data-Driven Stochastic Robust Optimization: General Computational Framework and Algorithm Leveraging Machine Learning for Optimization under Uncertainty in the Big Data Era

Chao Ning, Fengqi You*

Robert Frederick Smith School of Chemical and Biomolecular Engineering,

Cornell University, Ithaca, New York 14853, USA

## Abstract

A novel data-driven stochastic robust optimization (DDSRO) framework is proposed for optimization under uncertainty leveraging labeled multi-class uncertainty data. Uncertainty data in large datasets are often collected from various conditions, which are encoded by class labels. Machine learning methods including Dirichlet process mixture model and maximum likelihood estimation are employed for uncertainty modeling. A DDSRO framework is further proposed based on the data-driven uncertainty model through a bi-level optimization structure. The outer optimization problem follows a two-stage stochastic programming approach to optimize the expected objective across different data classes; adaptive robust optimization is nested as the inner problem to ensure the robustness of the solution while maintaining computational tractability. A decomposition-based algorithm is further developed to solve the resulting multi-level optimization problem efficiently. Case studies on process network design and planning are presented to demonstrate the applicability of the proposed framework and algorithm.

*Corresponding author. Phone: (607) 255-1162; Fax: (607) 255-9166; E-mail: fengqi.you@cornell.edu

# 1. Introduction

Labeled multi-class data are ubiquitous in a variety of areas and disciplines, such as process monitoring [1], multimedia studies [2, 3], and machine learning [4]. For example, process data are labeled with the operating modes of chemical plants [1], and text documents are tagged with labels to indicate their topics [5]. Due to the massive amount of available uncertainty data (realizations of uncertain parameters) and dramatic progress in big data analytics [6], data-driven optimization emerges as a promising paradigm for decision making under uncertainty [7-14]. Most existing data-driven optimization methods are restricted to unlabeled uncertainty data. However, recent work has revealed the significance of leveraging labeled uncertainty data in decision-making under uncertainty [15]. Applying the existing methods to labeled uncertainty data cannot take advantage of useful information embedded in labels. Consequently, novel modeling frameworks and computational algorithms are needed to leverage labeled multi-class uncertainty data in data-driven decision making under uncertainty.

Optimization under uncertainty has attracted tremendous attention from both academia and industry [16-21]. Uncertain parameters, if not being accounted for, could render the solution of an optimization problem suboptimal or even infeasible [22]. To this end, a plethora of mathematical programming techniques, such as stochastic programming and robust optimization [23-25], have been proposed for decision making under uncertainty. These techniques have their respective strengths and weaknesses, which lead to different application scopes [26-41]. Stochastic programming focuses on the expected performance of a solution by leveraging the scenarios of uncertainty realization and their probability distribution [42-44]. However, this approach requires accurate information on the probability distribution, and the resulting optimization problem could become computationally challenging as the number of scenarios increases [40, 45, 46]. Robust optimization provides an alternative approach that does not require accurate knowledge on probability distributions of uncertain parameters [47-51]. It has attracted considerable interest owing to the merit of feasibility guarantee and computational tractability [50]. Nevertheless, robust optimization does not take advantage of available probability distribution information, and its solution usually suffers from the conservatism issue. The state-of-the-art approaches for optimization under uncertainty leverage the synergy of different optimization methods to inherit their corresponding strengths and complement respective weaknesses [52-54]. However, these approaches do not take advantage of the recent advances in machine learning and big data analytics to leverage uncertainty data for optimization under uncertainty. Therefore, the research objective of our work is to propose a novel data-driven decision-making framework that organically integrates machine learning methods for labeled uncertainty data with the state-of-the-art optimization approach.

There are a number of research challenges towards a general data-driven stochastic robust optimization (DDSRO) framework. First, uncertainty data are often collected from a wide spectrum of conditions, which are indicated by categorical labels. For example, uncertain power generation data of a solar farm are labeled with weather conditions, such as cloudy, sunny and rainy [55]. The occurrence probability of each weather condition is known in the data collection process, and is typically embedded in the datasets. Moreover, the uncertainty data with the same label could exhibit very complicated distribution, which is impossible to fit into some commonly-known probability distributions. Therefore, a major research challenge is how to develop machine learning methods to accurately extract useful information from labeled multi-class uncertainty data for constructing the uncertainty model in the data-driven optimization framework. Second, we are confronted with the challenge of how to integrate different approaches for optimization under uncertainty in a holistic framework to leverage their respective advantages. The third challenge is to develop an efficient computational algorithm for the resulting large-scale multi-level optimization problem, which cannot be solved directly by any off-the-shelf optimization solvers.

This paper proposes a novel DDSRO framework that leverages machine learning for decision making under uncertainty. In large datasets, uncertainty data are often attached with labels to indicate their data classes [6]. The proposed framework takes advantage of machine learning methods to accurately and truthfully extract uncertainty information, including probability distributions of data classes and uncertainty sets. The probability distribution of data classes is learned from labeled uncertainty data through maximum likelihood estimation for the multinoulli distribution [4, 56]. To accurately capture the structure and complexity of uncertainty data, a group of Dirichlet process mixture models are employed to construct uncertainty sets with a variational inference algorithm [57, 58]. These two pieces of uncertainty information are incorporated into the DDSRO framework through a bi-level optimization structure. Two-stage stochastic programming is nested in the outer problem to optimize the expected objective over categorical data classes; adaptive robust optimization (ARO) is nested as the inner problem to hedge against uncertainty and ensure computational tractability. Estimating a categorical distribution is much more computationally tractable than estimating a joint probability distribution of high-dimensional uncertainties [4]. Therefore, the outer optimization problem follows a two-stage stochastic programming structure to take advantage of the available probability information on data classes. Robust optimization, which uses an uncertainty set instead of accurate knowledge on probability distributions, is more suitable to be nested as the inner problem to tackle high-dimensional uncertainties. The DDSRO problem is formulated as a multi-level mixed-integer linear program (MILP). We further develop a decomposition-based algorithm to efficiently solve the resulting multi-level MILP problem. To demonstrate the advantages of the proposed framework, two case studies on design and planning of process networks under uncertainty are presented.

The major novelties of this paper are summarized as follows.

- A novel uncertainty modeling framework based on machine learning methods that accurately extract different types of uncertainty information for optimization under uncertainty;
- A general data-driven decision making under uncertainty framework that combines machine learning with mathematical programming and that leverages the strengths of both stochastic programming and robust optimization;
- An efficient decomposition-based algorithm to solve the resulting data-driven stochastic robust four-level MILP problems.

The rest of this paper is organized as follows. Section 2 proposes the general DDSRO framework, which includes data-driven uncertainty model, optimization model and computational algorithm. A motivating example is presented in Section 3. It is followed by case studies on process network design and planning under uncertainty in Section 4. Conclusions are drawn in Section 5.

## 2. Data-Driven Stochastic Robust Optimization Framework

In this section, we first provide a brief introduction to stochastic robust optimization. The proposed data-driven uncertainty model and the DDSRO framework are presented next. A decomposition-based algorithm is further developed to efficiently solve the resulting multi-level optimization problem.

## 2.1. Stochastic robust optimization

The stochastic robust optimization framework combines two-stage stochastic programming and ARO [53]. In this subsection, we first analyze the strengths and weaknesses of two-stage stochastic programming and two-stage ARO. They are then integrated in a novel form to leverage their corresponding advantages and complement respective shortcomings.

A general two-stage stochastic MILP in its compact form is given as follows [23].

$$\begin{aligned} \min_{\mathbf{x}} \ & \mathbf{c}^{\mathrm{T}}\mathbf{x} + \mathbb{E}_{\mathbf{u}}\left[Q(\mathbf{x},\mathbf{u})\right] \\ \text{s.t.} \ & \mathbf{A}\mathbf{x} \geq \mathbf{d} \\ & \mathbf{x} \in R_+^{n_1} \times Z^{n_2} \end{aligned} \tag{1}$$

The recourse function $Q(\mathbf{x}, \mathbf{u})$ is defined as follows.

$$\begin{aligned} Q(\mathbf{x},\mathbf{u}) = \min_{\mathbf{y}} \ & \mathbf{b}^{\mathrm{T}}\mathbf{y} \\ \text{s.t.} \ & \mathbf{W}\mathbf{y} \geq \mathbf{h} - \mathbf{T}\mathbf{x} - \mathbf{M}\mathbf{u} \\ & \mathbf{y} \in R_+^{n_3} \end{aligned} \tag{2}$$

where $\mathbf{x}$ is first-stage decisions made "here-and-now" before the uncertainty $\mathbf{u}$ is realized, while the second-stage decisions or recourse decisions $\mathbf{y}$ are postponed in a "wait-and-see" manner after the uncertainties are revealed. $\mathbf{x}$ can include both continuous and integer variables, while $\mathbf{y}$ only includes continuous variables. The objective of (1) includes two parts: the first-stage objective $\mathbf{c}^{\mathrm{T}}\mathbf{x}$ and the expectation of the second-stage objective $\mathbf{b}^{\mathrm{T}}\mathbf{y}$. The constraints associated with the first-stage decisions are $\mathbf{A}\mathbf{x} \geq \mathbf{d},\ \mathbf{x} \in R_+^{n_1} \times Z^{n_2}$, and the constraints of the second-stage decisions are $\mathbf{W}\mathbf{y} \geq \mathbf{h} - \mathbf{T}\mathbf{x} - \mathbf{M}\mathbf{u},\ \mathbf{y} \in R_+^{n_3}$.

Two-stage stochastic programming makes full use of probability distribution information, and aims to find a solution that performs well on average under all scenarios. However, an accurate joint probability distribution of high-dimensional uncertainty $\mathbf{u}$ is required to calculate the expectation term in (1). Besides, the resulting two-stage stochastic programming problem could become computationally challenging as the number of scenarios increases [45].

Another popular approach for optimization under uncertainty is two-stage ARO, which hedges against the worst-case uncertainty realization within an uncertainty set. A general two-stage ARO problem in its compact form is given as follows [59].

$$
\begin{aligned}
&\min_{\mathbf{x}} \ \mathbf{c}^{\mathrm{T}}\mathbf{x} + \max_{\mathbf{u}\in U} \ \min_{\mathbf{y}\in\Omega(\mathbf{x},\mathbf{u})} \ \mathbf{b}^{\mathrm{T}}\mathbf{y} \\
&\text{s.t.} \quad \mathbf{A}\mathbf{x} \ge \mathbf{d}, \ \ \mathbf{x} \in R_{+}^{n_1} \times Z^{n_2} \\
&\qquad \Omega(\mathbf{x},\mathbf{u}) = \left\{ \mathbf{y} \in R_{+}^{n_3} : \mathbf{W}\mathbf{y} \ge \mathbf{h} - \mathbf{T}\mathbf{x} - \mathbf{M}\mathbf{u} \right\}
\end{aligned}
\tag{3}
$$

where $\mathbf{x}$ is the vector of first-stage decision variables, and $\mathbf{y}$ represents the vector of second-stage decisions. Note that $\mathbf{x}$ includes both continuous and integer variables, and $\mathbf{y}$ is the vector of continuous recourse variables. $U$ is an uncertainty set that characterizes the region in which uncertainty realization resides. Note that the two-stage ARO problem in (3) is a tri-level optimization problem.

The stochastic robust optimization framework is able to organically integrate the two-stage stochastic programming approach with the ARO method and leverage their strengths. A general form of the stochastic robust MILP is given by [53],

$$
\begin{aligned}
&\min_{\mathbf{x}} \ \mathbf{c}^{\mathrm{T}}\mathbf{x} + \mathbb{E}_{\sigma\in\Pi}\left[ \overline{C}_{\sigma}(\mathbf{x}) \right] \\
&\text{s.t.} \quad \mathbf{A}\mathbf{x} \ge \mathbf{d} \\
&\qquad \mathbf{x} \in R_{+}^{n_1} \times Z^{n_2} \\
&\qquad \left. \begin{cases} \overline{C}_{\sigma}(\mathbf{x}) = \max\limits_{\mathbf{u}\in U_{\sigma}} \ \min\limits_{\mathbf{y}_{\sigma}} \ \mathbf{b}^{\mathrm{T}}\mathbf{y}_{\sigma} \\ \qquad \text{s.t.} \quad \mathbf{W}\mathbf{y}_{\sigma} \ge \mathbf{h} - \mathbf{T}\mathbf{x} - \mathbf{M}\mathbf{u} \\ \qquad\qquad \mathbf{y}_{\sigma} \in R_{+}^{n_3} \end{cases} \right\}, \forall \sigma \in \Pi
\end{aligned}
\tag{4}
$$

where $\sigma$ is an uncertain scenario that influences the uncertainty set, and $\Pi$ is a set of the scenarios. Two-stage stochastic programming approach is nested in the outer problem to optimize the expectation of objectives for different scenarios, and robust optimization is nested inside to hedge against the worst-case uncertainty realization. Stochastic robust optimization is capable of handling different types of uncertainties in a holistic framework.

In stochastic programming, an accurate joint probability distribution of high-dimensional uncertainty $\mathbf{u}$ is required. However, from a statistical inference perspective, estimating a joint probability distribution of high-dimensional uncertainty is far more challenging than estimating accurate probability distributions of individual scenarios [4]. Therefore, it is better to use a coarse-grained uncertainty set to describe high-dimensional uncertainty $\mathbf{u}$, and to use fine-grained probability distributions to model scenarios. As a result, two-stage stochastic programming approach is nested outside to leverage the available probability distributions. Robust optimization is more suitable to be nested inside for computational tractability.

The stochastic robust optimization method typically assumes the uncertainty probability distributions and uncertainty sets are given and known a priori, rather than deriving them from the uncertainty data. However, the predefined probability distribution and manually constructed uncertainty sets might not correctly capture the fine-grained information from the available uncertainty data. Thus, we propose a novel data-driven uncertainty modeling framework for stochastic robust optimization with labeled multi-class uncertainty data to fill this knowledge gap.

## 2.2. Data-driven uncertainty modeling

In this subsection, we propose a data-driven uncertainty modeling framework based on machine learning techniques to seamlessly integrate data-driven system and model-based system in the proposed decision-making framework. This uncertainty modeling framework includes two parts, i.e. probability distribution estimation for data classes, and uncertainty set construction from uncertainty data in the same data class. For the first part, the probability distributions of data classes are learned from the label information in multi-class uncertainty data through maximum likelihood estimation. For the second part, a group of Dirichlet process mixture models is employed to construct uncertainty sets with a variational inference algorithm [58].

We consider the multi-class uncertainty data with labels $\left\{\mathbf{u}^{(i)}, c^{(i)}\right\}_{i=1}^{L}$, where $\mathbf{u}^{(i)}$ is $i$ th uncertainty data point, $c^{(i)}$ is its corresponding label, and $L$ is the total number of uncertainty data. Although the labeled uncertainty data is of practical relevance [6, 15], existing literature on data-driven optimization under uncertainty are restricted to unlabeled uncertainty data $\left\{\mathbf{u}^{(i)}\right\}_{i=1}^{L}$ [8, 12].

The uncertainty information on the probability of different data classes can be extracted from labeled uncertainty data by leveraging their label information. The occurrence probabilities of data classes are modeled with a multinoulli distribution. The probability of each data class can be calculated through maximum likelihood estimation, as given by [4, 56],

$$p_s = \frac{\sum_i \mathbb{I}\left(c^{(i)} = s\right)}{L} \tag{5}$$

where $p_s$ represents the occurrence probability of data class $s$, $c^{(i)}$ is the label associated with the $i$ th uncertainty data point, $c^{(i)}=s$ indicates that the $i$ th uncertainty data point is from data class $s$, and the indicator function is defined as follows.

$$\mathbb{I}\left(c^{(i)}=s\right)=\begin{cases}1 & \text{if } c^{(i)}=s\\ 0 & \text{else}\end{cases} \tag{6}$$

As will be shown in the next subsection, the extracted probability distribution information of data classes is incorporated into the two-stage stochastic programming framework nested as the outer optimization problem of DDSRO.

The second part of the uncertainty modeling is the construction of data-driven uncertainty sets from uncertainty data in the same data class. The information on uncertainty sets is incorporated into robust optimization nested as the inner optimization problem of DDSRO. To handle the labeled multi-class uncertainty data, a group of Dirichlet process mixture models is employed, and uncertainty data with the same label are modeled using one separate Dirichlet process mixture model [58]. The details of Dirichlet process mixture model are given in Appendix A.

Based on the extracted information from Dirichlet process mixture model, we construct a data-driven uncertainty set for data class $s$ using both $l_1$ and $l_\infty$ norms [8], as shown below,

$$U_s=\bigcup_{i:\gamma_{s,i}\geq\gamma^*}U_{s,i}=U_{s,1}\bigcup U_{s,2}\cdots\bigcup U_{s,m(s)} \tag{7}$$

$$U_{s,i}=\left\{\mathbf{u}\,\middle|\,\mathbf{u}=\boldsymbol{\mu}_{s,i}+\kappa_{s,i}\boldsymbol{\Psi}_{s,i}^{1/2}\Lambda_{s,i}\,\mathbf{z},\ \ \|\mathbf{z}\|_\infty\leq 1,\ \ \|\mathbf{z}\|_1\leq\Phi_{s,i}\right\} \tag{8}$$

where $U_s$ is the data-driven uncertainty set for data class $s$, $U_{s,i}$ is the basic uncertainty set corresponding to the $i$ th component of data class $s$, $\Lambda_{s,i}$ is a scaling factor, $\gamma_{s,i}$ is the weight of the $i$ th component of data class $s$, $\gamma_{s,i}=\dfrac{\tau_{s,i}}{\tau_{s,i}+\nu_{s,i}}\prod_{j=1}^{i-1}\dfrac{\nu_{s,j}}{\tau_{s,j}+\nu_{s,j}}$ and $\gamma_{s,M}=1-\sum_{i=1}^{M-1}\gamma_{s,i}$. The weight $\gamma_{s,i}$ indicates the probability of the corresponding component. $\gamma^*$ is a threshold value. $m(s)$ is the total number of components for data class $s$. $\kappa_{s,i}=\sqrt{\dfrac{\lambda_{s,i}+1}{\lambda_{s,i}\left(\omega_{s,i}+1-\dim(\mathbf{u})\right)}}$ [57], $\tau_{s,i}$, $\nu_{s,i}$, $\boldsymbol{\mu}_{s,i}$, $\lambda_{s,i}$, $\omega_{s,i}$, $\boldsymbol{\psi}_{s,i}$ are the inference results of the $i$ th component learned from uncertainty data of data class $s$, and $\Phi_{s,i}$ is its uncertainty budget. It is worth noting that the number of basic uncertainty sets $m(s)$ for each data class varies depending on the structure and complexity of the uncertainty data.

## 2.3. Data-driven stochastic robust optimization model

The proposed DDSRO framework is shown as follows.

$$
\begin{aligned}
&\min_{\mathbf{x}} \ \mathbf{c}^{\mathrm{T}}\mathbf{x} + \mathbb{E}_{s\in\Xi}\left[ \max_{\mathbf{u}\in U_{s,1}\bigcup U_{s,2}\cdots\bigcup U_{s,m(s)}} \ \min_{\mathbf{y}_s\in\Omega(\mathbf{x},\mathbf{u})} \mathbf{b}^{\mathrm{T}}\mathbf{y}_s \right] \\
&\text{s.t.} \quad \mathbf{A}\mathbf{x} \ge \mathbf{d}, \quad \mathbf{x}\in R_+^{n_1}\times Z^{n_2} \\
&\qquad U_{s,i} = \left\{ \mathbf{u} \,\middle|\, \mathbf{u} = \boldsymbol{\mu}_{s,i} + \kappa_{s,i}\boldsymbol{\Psi}_{s,i}^{1/2}\Lambda_{s,i}\,\mathbf{z}, \ \ \|\mathbf{z}\|_\infty \le 1, \ \ \|\mathbf{z}\|_1 \le \Phi_{s,i} \right\} \\
&\qquad \Omega(\mathbf{x},\mathbf{u}) = \left\{ \mathbf{y}_s \in R_+^{n_3} : \mathbf{W}\mathbf{y}_s \ge \mathbf{h} - \mathbf{T}\mathbf{x} - \mathbf{M}\mathbf{u} \right\}
\end{aligned} \tag{9}
$$

where $\Xi = \{1,2,\ldots,C\}$ is the set of data classes, $C$ is the total number of data classes, $m(s)$ is the total number of mixture components for data class $s$. Each data class corresponds to a scenario in the two-stage stochastic program nested outside the DDSRO framework. It is worth noting that the DDSRO problem in (9) is a multi-level MILP that involves mixed-integer first-stage decisions, and continuous recourse decisions.

In the DDSRO framework, the outer optimization problem follows a two-stage stochastic programming structure, and robust optimization is nested as the inner optimization problem to hedge against high-dimensional uncertainty $\mathbf{u}$. DDSRO aims to find a solution that performs well on average across all data classes by making full use of their occurrence probabilities. Instead of treating all the uncertainty data as a whole, the DDSRO framework explicitly accounts for multi-class uncertainty data by leveraging their label information. Stochastic robust optimization structure is employed because it is more suitable to leverage different types of uncertainty information in a holistic framework. Note that DDSRO is a general framework for data-driven optimization under uncertainty, while stochastic robust optimization methods assume uncertainty information is given *a priori* rather than learning it from uncertainty data.

There are two reasons that two-stage stochastic programming is nested as the outer optimization problem while robust optimization is nested as the inner optimization problem. First, the proposed DDSRO framework aims to balance all data classes instead of focusing only on the worst-case data class, and it also ensures the robustness of the solution. Second, estimating the probability distribution of data classes accurately is much more computationally tractable than estimating an accurate joint probability distribution [4]. Consequently, it is better to employ uncertainty set to characterize high-dimensional uncertain parameters in the inner optimization problem. On the other hand, stochastic programming is more suitable as the outer optimization problem to leverage the probability information of data classes.

The DDSRO framework leverages machine learning techniques to truthfully capture the structure and complexity of uncertainty data for decision making under uncertainty. It enjoys the computational tractability of robust optimization, while accounting for exact probability distribution of data class in the same way as stochastic programming approach.

Note that the proposed data-driven uncertainty set for each data class in (7) is a union of several basic uncertainty sets. Therefore, we reformulate the DDSRO model in (9) into (10), as given below.

$$\begin{aligned} \min_{\mathbf{x}} \ & \mathbf{c}^{\mathrm{T}}\mathbf{x} + \mathbb{E}_{s\in\Xi}\left[ \max_{i\in\{1,\ldots,m(s)\}} \max_{\mathbf{u}\in U_{s,i}} \min_{\mathbf{y}_s\in\Omega(\mathbf{x},\mathbf{u})} \mathbf{b}^{\mathrm{T}}\mathbf{y}_s \right] \\ \text{s.t.} \ & \mathbf{Ax} \geq \mathbf{d}, \quad \mathbf{x} \in R_+^{n_1} \times Z^{n_2} \\ & U_{s,i} = \left\{ \mathbf{u} \,\middle|\, \mathbf{u} = \boldsymbol{\mu}_{s,i} + \kappa_{s,i}\boldsymbol{\Psi}_{s,i}^{1/2}\Lambda_{s,i}\,\mathbf{z},\ \ \|\mathbf{z}\|_\infty \leq 1,\ \ \|\mathbf{z}\|_1 \leq \Phi_{s,i} \right\} \\ & \Omega(\mathbf{x},\mathbf{u}) = \left\{ \mathbf{y}_s \in R_+^{n_3} : \mathbf{W}\mathbf{y}_s \geq \mathbf{h} - \mathbf{Tx} - \mathbf{Mu} \right\} \end{aligned} \tag{10}$$

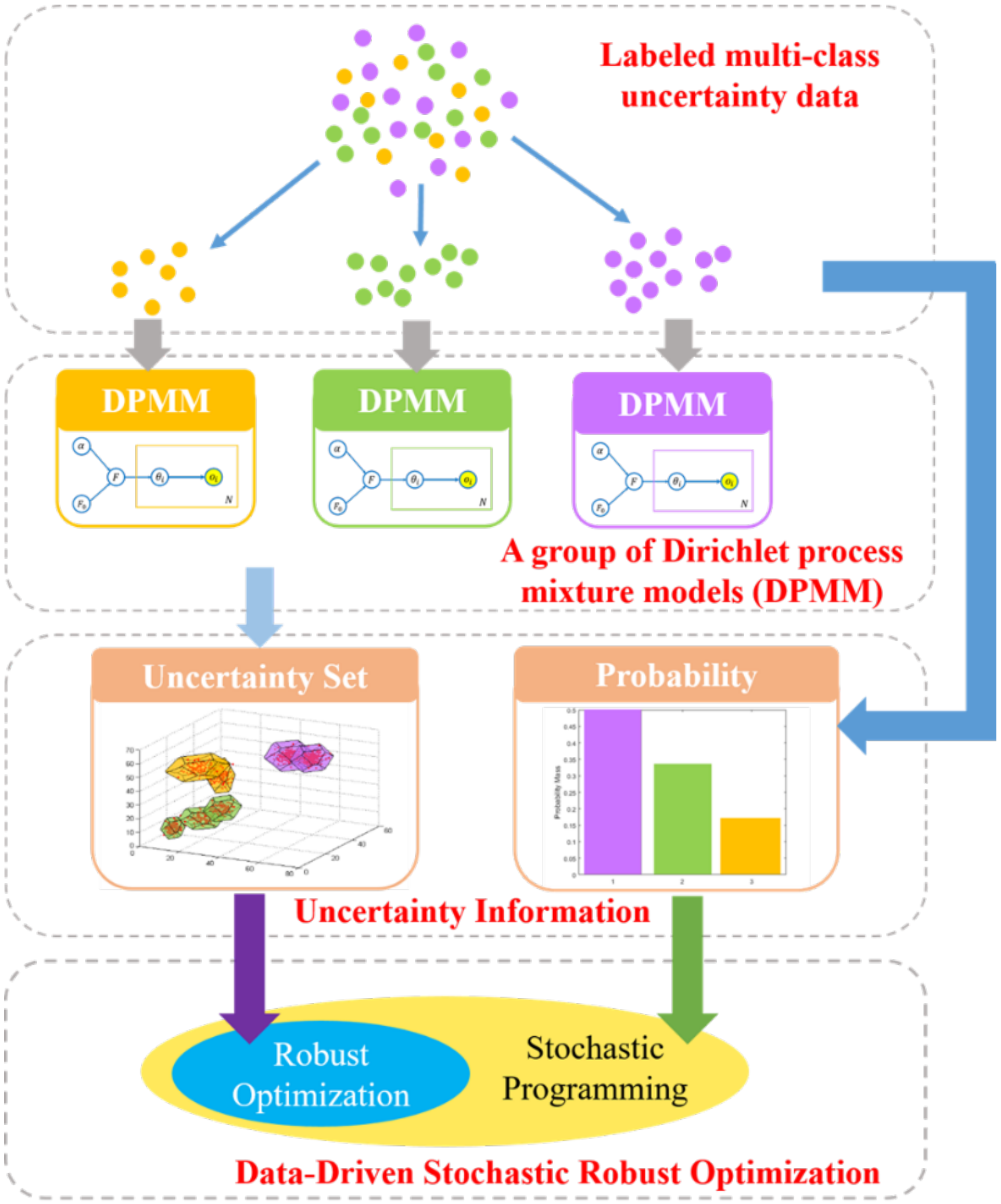

Fig. 1. An overview of the proposed DDSRO framework.

The DDSRO model explicitly accounts for the label information within multi-class uncertainty data, and balances different data classes by using the weighted sum based on their occurrence probabilities. Robust optimization is nested as the inner optimization problem to handle uncertainty **u**, thus providing a good balance between solution quality and computational tractability. Moreover, multiple basic uncertainty sets $U_{s,i}$ are unified to capture the structure and complexity of uncertainty data, and this scheme leads to the inner max-max optimization problem in (15). An overview of the proposed DDSRO framework for labeled multi-class uncertainty data is shown in Fig. 1. Uncertainty data points are depicted using circles with different colors to indicate its associated data class, and they are treated separately using a group of Dirichlet process mixture models. The probability distribution of data classes can be inferred from uncertainty data using eq. (5). The proposed framework leverages the strengths of stochastic programming and robust optimization, and uses the uncertainty information embedded in uncertainty data.

The resulting DDSRO problem has a multi-level optimization structure that cannot be solved by any general-purpose optimization solvers directly. An efficient computational algorithm is further proposed in the next subsection to address this computational challenge.

## 2.4. The decomposition-based algorithm

To address computational challenge of solving the DDSRO problem, we extend the algorithm proposed in [8]. The main idea is to first decompose the multi-level MILP shown in (15) into a master problem and a number of groups of sub-problems, where each group corresponds to a data class. Moreover, there are a number of sub-problems in each group, and each sub-problem corresponds to a component in the Dirichlet process mixture model.

The proposed decomposition-based algorithm iteratively solves the master problem and the sub-problems, and adds cuts to the master problem after iteration, until the relative optimality gap is reduced to the predefined tolerance $\zeta$. The pseudocode of the algorithm is given in Fig. 2, where $C$ represents the total number of data classes. The algorithm is guaranteed to terminate within finite iterations due to the finite number of extreme points of basic uncertainty sets [53]. The details of the decomposition-based algorithm, including the formulations of master problem (**MP**) and sub-problem ($\mathbf{SUP}_{s,i}$), are given in Appendix B.

**Decomposition Algorithm**

Set $LB \leftarrow -\infty$, $UB \leftarrow +\infty$, $k \leftarrow 0$, and $\zeta$;
**while** $\left|\frac{UB-LB}{UB}\right| > \zeta$ **do**
  Solve (**MP**) to obtain $\mathbf{x}_{k+1}^*, \eta_{k+1}^*, \mathbf{y}_s^{1*}, \ldots, \mathbf{y}_s^{k*}$;
  Update $LB \leftarrow \mathbf{c}^{\mathrm{T}}\mathbf{x}_{k+1}^* + \eta_{k+1}^*$;
  **for** $s$=1 to $C$ **do**
    **for** $i = 1$ to $m(s)$ **do**
      Solve (**SUP*s,i***) to obtain $\mathbf{u}_{s,i}^{k+1}$ and $Q_{s,i}\left(\mathbf{x}_{k+1}^*\right)$;
    **end**
    $i^* \leftarrow \arg\max_i Q_{s,i}\left(\mathbf{x}_{k+1}^*\right)$;
    $\mathbf{u}_s^{k+1} \leftarrow \mathbf{u}_{s,i^*}^{k+1}$ and $Q_s\left(\mathbf{x}_{k+1}^*\right) \leftarrow Q_{s,i^*}\left(\mathbf{x}_{k+1}^*\right)$;
  **end**
  Update $UB \leftarrow \min\left\{UB, \mathbf{c}^{\mathrm{T}}\mathbf{x}_{k+1}^* + \sum_{s=1} p_s \cdot Q_s\left(\mathbf{x}_{k+1}^*\right)\right\}$;
  Create second-stage variables $\mathbf{y}_s^{k+1}$ and add cuts $\eta \geq \sum_s p_s\left(\mathbf{b}^{\mathrm{T}}\mathbf{y}_s^{k+1}\right)$ and $\mathbf{Tx} + \mathbf{Wy}_s^{k+1} \geq \mathbf{h} - \mathbf{Mu}_s^{k+1}$ to (**MP**);
  $k \leftarrow k+1$;
**end**
**return** $UB$;

Fig. 2. The pseudocode of the decomposition-based algorithm.

## 2.5. Relationship with the existing data-driven adaptive nested robust optimization framework

In this subsection, we discuss the relationship between the DDSRO framework proposed in this work and the data-driven adaptive nested robust optimization (DDANRO) framework proposed earlier [8].

The DDANRO framework in its general form has a four-level optimization structure, as given below.

$$
\begin{aligned}
&\min_{\mathbf{x}} \ \mathbf{c}^{\mathrm{T}}\mathbf{x} + \max_{i\in\{1,\dots,m\}} \max_{\mathbf{u}\in U_i} \min_{\mathbf{y}\in\Omega(\mathbf{x},\mathbf{u})} \mathbf{b}^{\mathrm{T}}\mathbf{y} \\
&\text{s.t.} \quad \mathbf{Ax} \ge \mathbf{d}, \quad \mathbf{x} \in R_+^{n_1} \times Z^{n_2} \\
&\qquad U_i = \left\{ \mathbf{u} \,\middle|\, \mathbf{u} = \boldsymbol{\mu}_i + \kappa_i \boldsymbol{\Psi}_i^{1/2} \Lambda_i \, \mathbf{z}, \ \ \|\mathbf{z}\|_\infty \le 1, \ \|\mathbf{z}\|_1 \le \Delta_i \right\} \\
&\qquad \Omega(\mathbf{x},\mathbf{u}) = \left\{ \mathbf{y} \in R_+^{n_3} : \mathbf{Wy} \ge \mathbf{h} - \mathbf{Tx} - \mathbf{Mu} \right\}
\end{aligned}
\tag{11}
$$

It is worth noting that the DDANRO framework treats uncertainty data homogeneously without considering the label information [8]. DDSRO greatly enhances the capabilities of DDANRO, and expands its application scope to more complex uncertainty data structures. There are some similarities between DDSRO and DDANRO. For example, they both employ data-driven uncertainty sets, and have multi-level optimization structures. Moreover, DDANRO can be considered as a special case of DDSRO when there is one only one class of uncertainty data. Specifically, if there is only one data class, we can simplify the expectation term as a single term in (10), such that the DDSRO model shown in (10) reduces to the DDANRO model in (11). The DDANRO framework is more appropriate to handle single-class or unlabeled uncertainty data, while the proposed DDSRO framework, on the other hand, is suitable for handling labeled multi-class uncertainty data systematically. Therefore, the DDSRO framework proposed in this paper could be considered as a generalization of the existing DDANRO approach.

## 3. Motivating Example

In this section, a motivating example is first presented to illustrate different characteristics of optimization methods, and to show that accounting for label information in optimization problem is critical.

The deterministic formulation of the motivating example is shown in (12).

$$
\begin{aligned}
&\min_{\mathbf{x},\mathbf{y}} \ 3x_1 + 5x_2 + 6x_3 + 6y_1 + 10y_2 + 12y_3 \\
&\text{s.t.} \quad x_1 + x_2 + x_3 \le 200 \\
&\qquad x_1 + y_1 \ge u_1 \\
&\qquad x_2 + y_2 \ge u_2 \\
&\qquad x_3 + y_3 \ge u_3 \\
&\qquad x_i, y_i \ge 0, \quad i = 1,2,3
\end{aligned}
\tag{12}
$$

where $x_1$, $x_2$, $x_3$, $y_1$, $y_2$ and $y_3$ are decision variables, $u_1$, $u_2$ and $u_3$ are parameters. In the deterministic optimization problem, $u_1$, $u_2$ and $u_3$ are assumed to be known exactly.

Under the uncertain environment, the parameters $u_1$, $u_2$ and $u_3$ are subject to uncertainty. In a data-driven optimization setting, what decision makers know about these uncertain parameters are their realizations, also known as uncertainty data [60]. The scatter plot of 1,000 labeled uncertainty data is shown in Fig. 1, where the points represent uncertainty realizations of $u_1$, $u_2$ and $u_3$. There are 4 labels within uncertainty data, which means that there are 4 data classes in total. Note that uncertainty data points from different data classes are indicated with disparate shapes in Fig. 1. Moreover, uncertainty data in the same data class could exhibit complex characteristics, such as correlation, asymmetry and multimode.

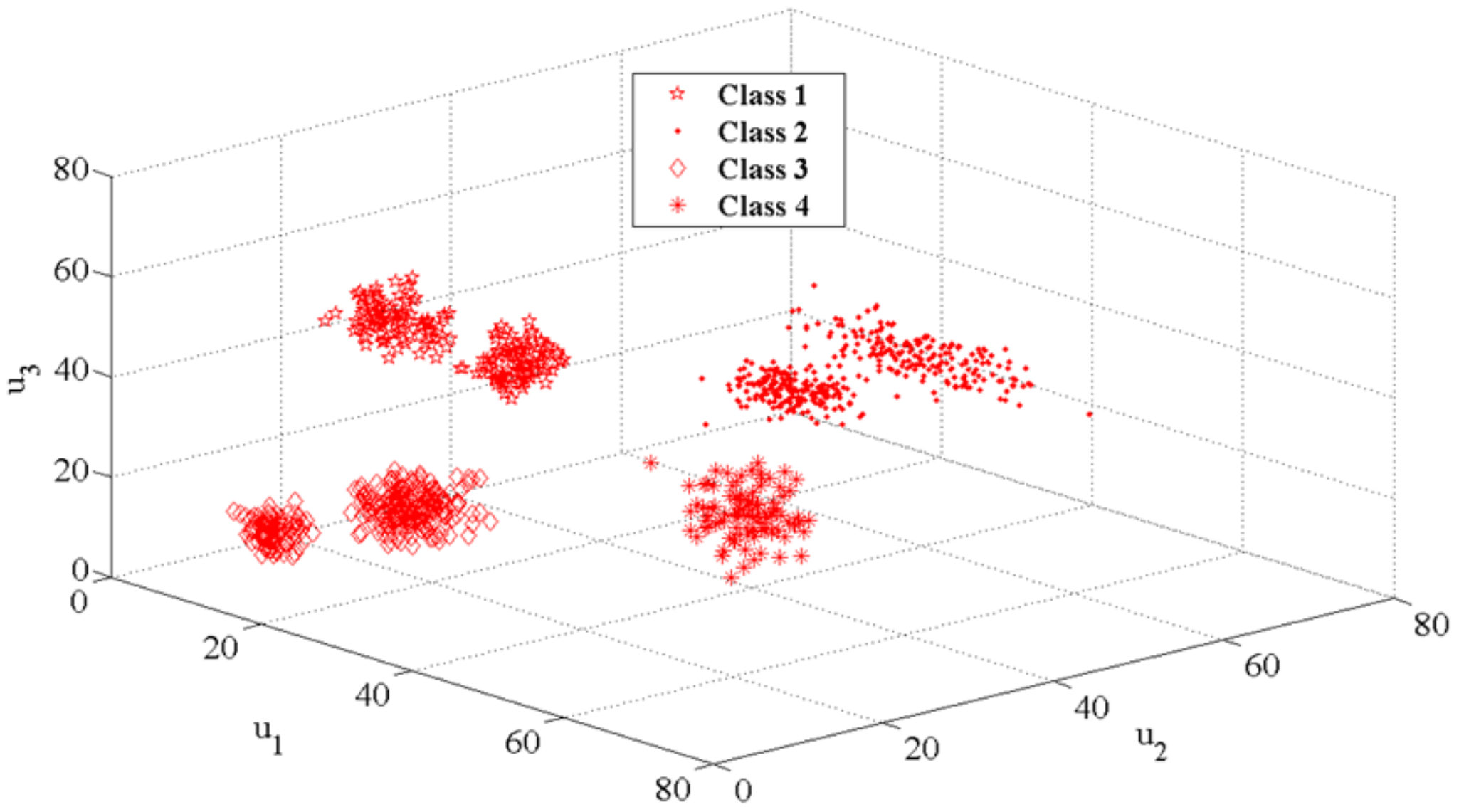

Fig. 3. The scatter plot of labeled uncertainty data in the motivating example.

One conventional way to hedging against uncertainty is the two-stage stochastic programming approach [23]. However, it is computationally challenging to fit the uncertainty data points in Fig. 3 into a reasonable probability distribution. If a scenario-based stochastic programming approach is applied to this problem, each data point in Fig. 3 corresponds to a scenario. There are 1,000 scenarios in total, and the probability of each scenario is given as 0.001. The scenario-based two-stage stochastic programming model is shown as follows.

$$
\begin{aligned}
\min_{\mathbf{x},\mathbf{y}_\omega} \ & 3x_1 + 5x_2 + 6x_3 + \sum_{\omega\in\Theta} p_\omega \cdot \left[ 6\mathrm{y}_{\omega,1} + 10y_{\omega,2} + 12y_{\omega,3} \right] \\
\text{s.t.} \ & x_1 + x_2 + x_3 \le 200 \\
& x_1 + y_{\omega,1} \ge u_{\omega,1} \quad \forall \omega \\
& x_2 + y_{\omega,2} \ge u_{\omega,2} \quad \forall \omega \\
& x_3 + y_{\omega,3} \ge u_{\omega,3} \quad \forall \omega \\
& x_i, y_{\omega,i} \ge 0, \qquad \forall \omega,\ i = 1,2,3
\end{aligned} \tag{13}
$$

where the notation Θ represents the set of scenarios, and $\omega$ is the scenario index. The probability of each scenario is $p_\omega$. $x_1$, $x_2$ and $x_3$ are the first-stage or "here-and-now" decision variables, $u_{\omega,1}$, $u_{\omega,2}$ and $u_{\omega,3}$ are the uncertainty realizations, and $y_{\omega,1}$, $y_{\omega,2}$ and $y_{\omega,3}$ are the second-stage or "wait-and-see" decision variables corresponding to scenario $\omega$. Note that the "here-and-now" decisions are made prior to the resolution of uncertainties, whereas "wait-and-see" decisions could be made after the uncertainties are revealed. The deterministic equivalent of this stochastic program will be a large-scale linear program with 3,003 variables and 6,004 constraints based on the 1,000 scenarios. We note that the size of the stochastic programming problem grows exponentially as the number of uncertainty data points increases. While decomposition-based optimization algorithms, such as multi-cut Benders decomposition algorithm, can take advantage of the problem structure and improve the computational efficiency of solving this problem [61], it remains intractable to handle problems with "big" data for uncertain parameters (e.g. those with millions of data points).

An alternative way to optimization under uncertainty is the two-stage ARO method [50], of which the problem size might be less sensitive to the amount of uncertainty data. The two-stage ARO model with a box uncertainty set is given as follows.

$$
\begin{aligned}
\min_{\mathbf{x}} \ & 3x_1 + 5x_2 + 6x_3 + \max_{\mathbf{u}\in U} \ \min_{\mathbf{y}} \ 6\mathrm{y}_1 + 10y_2 + 12y_3 \\
\text{s.t.} \ & x_1 + x_2 + x_3 \le 200 \\
& x_1 + y_1 \ge u_1 \\
& x_2 + y_2 \ge u_2 \\
& x_3 + y_3 \ge u_3 \\
& U = \left\{ \mathbf{u} \middle| u_i^L \le u_i \le u_i^U,\ i = 1,2,3 \right\} \\
& x_i, y_i \ge 0, \quad i = 1,2,3
\end{aligned} \tag{14}
$$

where $u_i^L$ and $u_i^U$ are the lower and upper bounds of uncertain parameter $u_i$, respectively. These two parameters can be readily derived from the uncertainty data in Fig. 3.

The numerical results are shown in Table 1. In the deterministic optimization method, $u_1$, $u_2$ and $u_3$ are set to be their mean values of the uncertainty data. Although the deterministic approach achieves a low objective value of 455.3 in the minimization problem, its optimal solution could become infeasible when the parameters $u_1$, $u_2$ and $u_3$ are subject to uncertainty. The two-stage stochastic programming approach focuses on the expected objective value, while the two-stage ARO method aims to minimize the worst-case objective value. Therefore, the objective value of the two-stage stochastic programming approach is lower than that of the two-stage ARO method. Compared with the two-stage stochastic programming approach, the two-stage ARO method is more computationally efficient.

A large amount of uncertainty data implies there is much uncertainty information. Nevertheless, the two-stage stochastic programming problem could become computationally intractable as the size of uncertainty data increases. Moreover, the stochastic programming approach could not well capture uncertainty data if we fit the uncertainty data into an inaccurate probability distribution, thereby resulting in poor-quality solutions. Consequently, it might be unsuitable to apply the conventional two-stage stochastic programming approach to the optimization problem where large amounts of uncertainty data are available. Despite its computational efficiency, the conventional two-stage ARO method only extracts coarse-grained uncertainty information on the bounds of uncertain parameters. Neglecting relevant uncertainty information makes the conventional ARO method unsuitable for the data-driven optimization setting either.

The DDANRO framework was recently proposed to overcome the drawback of the conventional ARO method [8] by integrating machine learning methods with ARO. The DDANRO model of this motivating example is given by

$$
\begin{aligned}
\min_{\mathbf{x}} \ & 3x_1 + 5x_2 + 6x_3 + \max_{j\in\{1,\ldots,m\}} \max_{\mathbf{u}\in U_j} \min_{\mathbf{y}} \ 6\mathrm{y}_1 + 10y_2 + 12y_3 \\
\text{s.t.} \ & x_1 + x_2 + x_3 \le 200 \\
& x_1 + y_1 \ge u_1 \\
& x_2 + y_2 \ge u_2 \\
& x_3 + y_3 \ge u_3 \\
& x_i, y_i \ge 0, \quad i = 1,2,3 \\
& U_j = \left\{ \mathbf{u} \,\middle|\, \mathbf{u} = \boldsymbol{\mu}_j + \kappa_j \boldsymbol{\Psi}_j^{1/2} \Lambda_j \, \mathbf{z}, \ \ \|\mathbf{z}\|_\infty \le 1, \ \ \|\mathbf{z}\|_1 \le \Delta_j \right\}
\end{aligned}
\tag{15}
$$

where $U_j$ is a basic uncertainty set, $\boldsymbol{\mu}_j$, $\kappa_j$, $\boldsymbol{\psi}_j$, and $\Lambda_j$ are the parameters derived from uncertainty data [8]. $\Delta_j$ is an uncertainty budget, and $m$ represents the total number of the basic uncertainty sets.

Nevertheless, one limitation of the DDANRO method is that it does not leverage the label information in Fig. 3 to infer the occurrence probability of each data class. This information derived from labels should inform the decision-making process. Besides, DDANRO does not leverage the strengths of stochastic programming for tackling uncertainty in optimization. These motivate us to propose the DDSRO framework in this paper as an extension of the DDANRO method for optimization under labeled multi-class uncertainty data.

The DDSRO model formulation of this motivating example is given in (16).

$$\begin{aligned} \min_{\mathbf{x}} \ & 3x_1 + 5x_2 + 6x_3 + \mathbb{E}_{s\in\Xi}\left[ \max_{\mathbf{u}\in U_s} \ \min_{\mathbf{y}} \ 6\mathrm{y}_1 + 10y_2 + 12y_3 \right] \\ \text{s.t.} \ & x_1 + x_2 + x_3 \le 200 \\ & x_1 + y_1 \ge u_1 \\ & x_2 + y_2 \ge u_2 \\ & x_3 + y_3 \ge u_3 \\ & x_i, y_i \ge 0, \quad i = 1,2,3 \end{aligned} \tag{16}$$

where the notation $\Xi$ represents the set of data classes, and $s$ is the corresponding element. The data-driven uncertainty set $U_s$ for data class $s$ is defined in (7).

Following the proposed data-driven uncertainty modeling framework in Section 2, the occurrence probabilities of different data classes are inferred from the label information embedded within uncertainty data using maximum likelihood estimation. The computational results show that the probabilities of data classes 1-4 are 0.2, 0.4, 0.3 and 0.1, respectively. Moreover, the uncertainty modeling results also show that there are two components in the Dirichlet process mixture models for data class 1-3, while there is only one component for data class 4. The occurrence probabilities of different data classes are listed in Fig. 4. The data-driven uncertainty sets with uncertainty budget $\Phi$=1.8 are also constructed for each data class, and are depicted using different colors in Fig. 4. From Fig. 4, we can see that the structure and complexity of uncertainty data are accurately and truthfully captured by the proposed data-driven uncertainty model. For example, the uncertainty set for data class 2 accurately captures the bimodal feature in uncertainty data, and the correlations among uncertain parameters $u_1$, $u_2$ and $u_3$. The computational results of the DDANRO and DDSRO methods are also provided in Table 1. By leveraging the label information, the DDSRO approach is less conservative than the DDANRO method. From Table 1, we can see that the objective value of DDANRO is 18.8% larger than that of DDSRO, while the computational times of both problems are comparable.

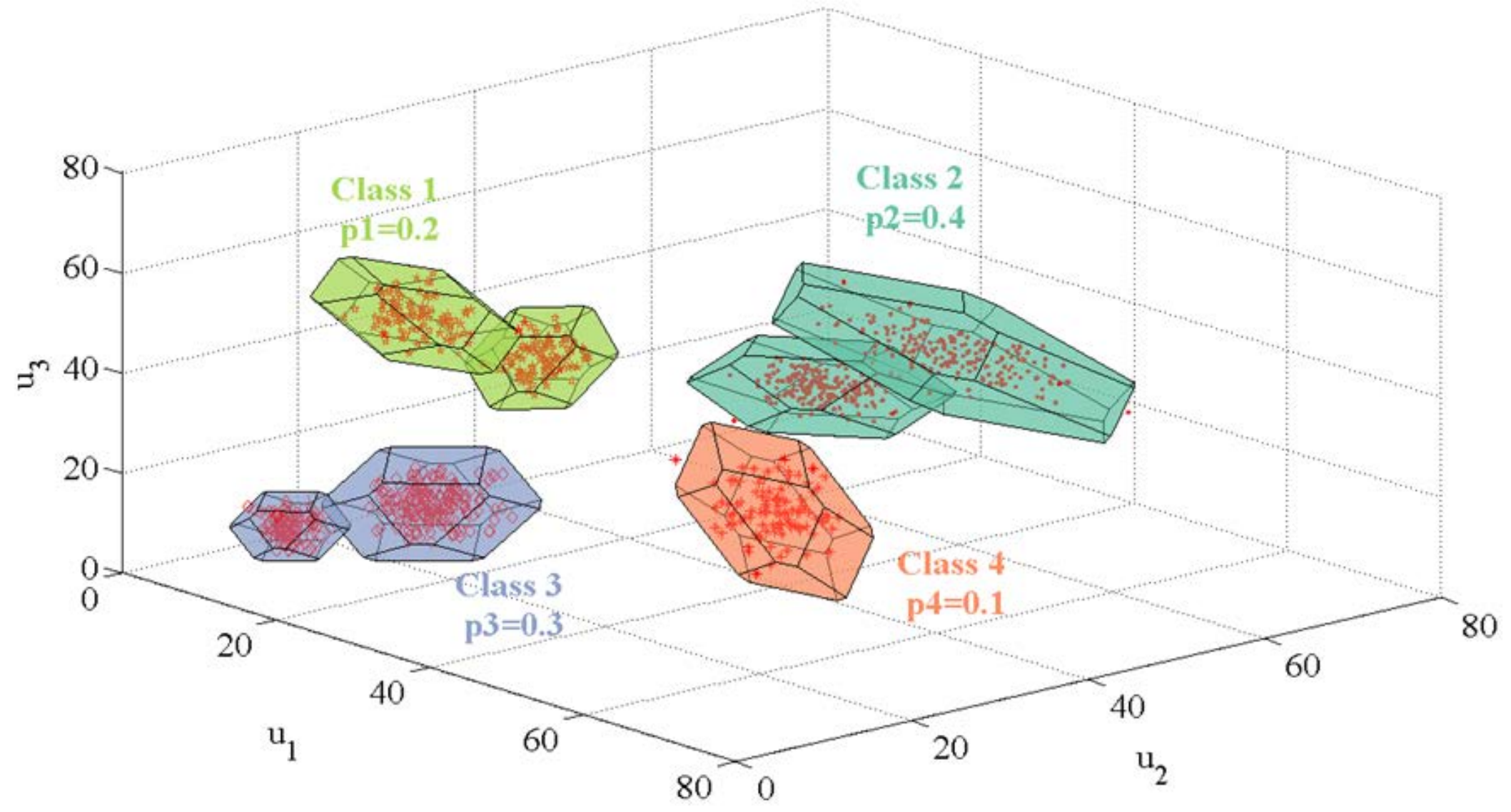

Fig. 4. The data-driven uncertainty modeling results for the four data classes in the motivating example.

Table 1. Computational results of different optimization methods in the motivating example.

| | Deterministic optimization | Two-stage stochastic programming | Two-stage ARO | DDANRO $\Delta$=1.8* | DDSRO $\Phi$=1.8# |
|---|---|---|---|---|---|
| Min.obj. | 455.3 | 646.4 | 964.0 | 944.6 | 795.4 |
| Iterations | N/A | 7 | 2 | 5 | 6 |
| Total CPU (s) | 0.1 | 477 | 1 | 9 | 12 |
| | $x_1 = 33.93$ | $x_1 = 38.43$ | $x_1 = 66.64$ | $x_1 = 55.48$ | $x_1 = 25.18$ |
| Decision **x** | $x_2 = 29.55$ | $x_2 = 27.55$ | $x_2 = 62.49$ | $x_2 = 52.56$ | $x_2 = 33.62$ |
| | $x_3 = 34.30$ | $x_3 = 39.92$ | $x_3 = 70.87$ | $x_3 = 67.60$ | $x_3 = 53.14$ |

* $\Delta$ is the data-driven uncertainty budget in the DDANRO approach.

# $\Phi$ is the data-driven uncertainty budget in the DDSRO approach.

# 4. Application: Data-Driven Stochastic Robust Planning of Chemical Process Networks under Uncertainty

The application of the proposed DDSRO framework to a multi-period process network planning problem is presented in this section. Large integrated chemical complexes consist of interconnected processes and various chemicals (see Fig. 9 for an example) [62, 63]. These interconnections allow the chemical production to make the most of different processes [64, 65]. The chemicals in the process network include raw materials, intermediates and final products. In the process network planning problem, purchase levels of raw materials, sales of products, capacity expansions, and production profiles of processes at each time period should be determined in order to maximize the net present value (NPV) during the entire planning horizon [66].

## 4.1. Data-driven stochastic robust planning of process networks model

The DDSRO model for process network planning under uncertainty is formulated as follows. The objective is to maximize the NPV, which is given in (17). The constraints include capacity expansion constraints (18)-(19), budget constraints (20)-(21), production level constraint (22), mass balance constraint (23), supply and demand constraints (24)-(25), non-negativity constraints (26)-(27), and integrity constraint (28). A list of indices/sets, parameters and variables is given in Nomenclature, where all the parameters are in lower-case symbols, and all the variables are denoted in upper-case symbols.

The objective function is defined by (17).

$$\max_{QE_{it},Y_{it}} -\sum_{i\in I}\sum_{t\in T}\left(c1_{it}\cdot QE_{it}+c2_{it}\cdot Y_{it}\right)+\mathbb{E}_{s\in\Xi}\left[\min_{du_{jt}\in U_s^{\mathrm{dem}},su_{jt}\in U_s^{\mathrm{sup}}}\ \max_{\substack{P_{sjt},Q_{sit},\\ SA_{sjt},W_{sit}}}\left(\sum_{j\in J}\sum_{t\in T}v_{jt}\cdot SA_{sjt}-\sum_{i\in I}\sum_{t\in T}c3_{it}\cdot W_{sit}-\sum_{j\in J}\sum_{t\in T}c4_{jt}\cdot P_{sjt}\right)\right] \tag{17}$$

where $QE_{it}$ is a decision variable for capacity expansion of process $i$ at period $t$, $Y_{it}$ is a binary decision variable to determine on whether process $i$ is expanded at period $t$, $SA_{sjt}$ is the amount of chemical $j$ sold to the market at period $t$ for data class $s$, $W_{sit}$ represents the operating level of process $i$ at period $t$ for data class $s$, and $P_{sjt}$ is a decision variable on the amount of chemical $j$ purchased at period $t$ for data class $s$. $c1_{it}$, $c2_{it}$, $c3_{it}$ and $c4_{jt}$ are the coefficients associated with variable investment cost, fixed investment cost, operating cost and purchase cost, respectively. $v_{jt}$ is the sale price of chemical $j$ in time period $t$.

Constraint (18) specifies the lower and upper bounds of the capacity expansions. Specifically, if $Y_{it}$=1, i.e. process *i* is selected to be expanded at period *t*, the expanded capacity should be within the range [$qe_{it}^L$, $qe_{it}^U$].

$$qe_{it}^{L} \cdot Y_{it} \le QE_{it} \le qe_{it}^{U} \cdot Y_{it}, \quad \forall i,t \tag{18}$$

Constraint (19) depicts the update of capacity of process *i*.

$$Q_{it} = Q_{it-1} + QE_{it}, \quad \forall i,t \tag{19}$$

where $Q_{it}$ is a decision variable for total capacity of process *i* at period *t*.

Constraints (20) and (21) enforce the largest number of capacity expansions for process and investment budget, respectively.

$$\sum_{t} Y_{it} \le ce_{i}, \quad \forall i \tag{20}$$

$$\sum_{i} \left( c1_{it} \cdot QE_{it} + c2_{it} \cdot Y_{it} \right) \le cb_{t}, \quad \forall t \tag{21}$$

where $ce_i$ is the largest possible expansion times of process *i*, and $cb_t$ is the investment budget for period *t*.

Constraint (22) specifies that the production level of a process cannot exceed its total capacity.

$$W_{sit} \le Q_{it}, \quad \forall s,i,t \tag{22}$$

Constraint (23) specifies the mass balance for chemicals.

$$P_{sjt} - \sum_{i} \kappa_{ij} \cdot W_{sit} - SA_{sjt} = 0, \quad \forall s,j,t \tag{23}$$

where $\kappa_{ij}$ represents the mass balance coefficient of chemical *j* in process *i*.

Constraint (24) specifies that the purchase amount of a feedstock cannot exceed its available amount. The sale amount of a product and its market demand are specified in Constraint (25).

$$P_{sjt} \le su_{jt}, \quad \forall s,j,t \tag{24}$$

$$SA_{sjt} \le du_{jt}, \quad \forall s,j,t \tag{25}$$

where $su_{jt}$ is the market availability and $du_{jt}$ is the demand of chemical *j* in time period *t*.

Constraints (26)-(27) enforces the non-negativity of continuous decisions.

$$QE_{it} \ge 0,\ Q_{it} \ge 0 \quad \forall i,t \tag{26}$$

$$P_{sjt},\ SA_{sjt},\ W_{sit} \ge 0, \quad \forall s,j,t \tag{27}$$

Constraint (28) ensures that $Y_{it}$ is a binary decision variable.

$$Y_{it} \in \{0,1\}, \quad \forall i,t \tag{28}$$

Uncertainty information for demand and supply is shown in (29)-(30).

$$U_s^{\text{dem}} = \bigcup_{k:\gamma_{s,k}^{dem} \ge \gamma^*} \left\{ \mathbf{du} \middle| \mathbf{du} = \boldsymbol{\mu}_{s,k}^{\text{dem}} + \kappa_{s,k}^{\text{dem}} \cdot \sqrt{\boldsymbol{\Psi}_{s,k}^{\text{dem}}} \cdot \Lambda_{s,k}^{\text{dem}} \cdot \mathbf{z},\ \ \|\mathbf{z}\|_\infty \le 1,\ \ \|\mathbf{z}\|_1 \le \Phi_{s,k}^{\text{dem}} \right\} \tag{29}$$

where $\boldsymbol{\mu}_{s,k}^{dem}$, $\kappa_{s,k}^{dem}$, $\boldsymbol{\Psi}_{s,k}^{dem}$ are the inference results learned from uncertain product demand data using the variational inference algorithm.

$$U_s^{\text{sup}} = \bigcup_{k:\gamma_{s,k}^{\text{sup}} \ge \gamma^*} \left\{ \mathbf{su} \middle| \mathbf{su} = \boldsymbol{\mu}_{s,k}^{\text{sup}} + s_{s,k}^{\text{sup}} \cdot \sqrt{\boldsymbol{\Psi}_{s,k}^{\text{sup}}} \cdot \Lambda_{s,k}^{\text{sup}} \cdot \mathbf{z},\ \ \|\mathbf{z}\|_\infty \le 1,\ \ \|\mathbf{z}\|_1 \le \Phi_{s,k}^{\text{sup}} \right\} \tag{30}$$

where $\boldsymbol{\mu}_{s,k}^{\text{sup}}$, $\kappa_{s,k}^{\text{sup}}$, $\boldsymbol{\Psi}_{s,k}^{\text{sup}}$ are the inference results learned from uncertain supply data employing the variational inference algorithm. The probabilities of different data classes are extracted using maximum likelihood estimation.

The resulting problem on strategic planning of process network under uncertainty is a stochastic robust MILP. The proposed computational algorithm is employed to efficiently solve the large-scale multi-level optimization problem. In the next two subsections, two case studies are presented to demonstrate the applicability and advantages of the proposed approach.

## 4.2. Case study 1

A small-scale case study is provided in this subsection. The process network, which is shown in Fig. 5, consists of five chemicals and three processes [63]. In Fig. 5, chemicals A-C represent raw materials, which can be either purchased from suppliers or produced by certain processes. For example, Chemical C can be either manufactured by Process 3 or purchased from a supplier. Chemicals D and E are final products, which are sold to the markets. In this case study, we consider 10 time periods over the 20-year planning horizon, and the duration of each time period is 2 years. It is assumed that processes do not have initial capacities, and they can be installed at the beginning of the planning horizon.

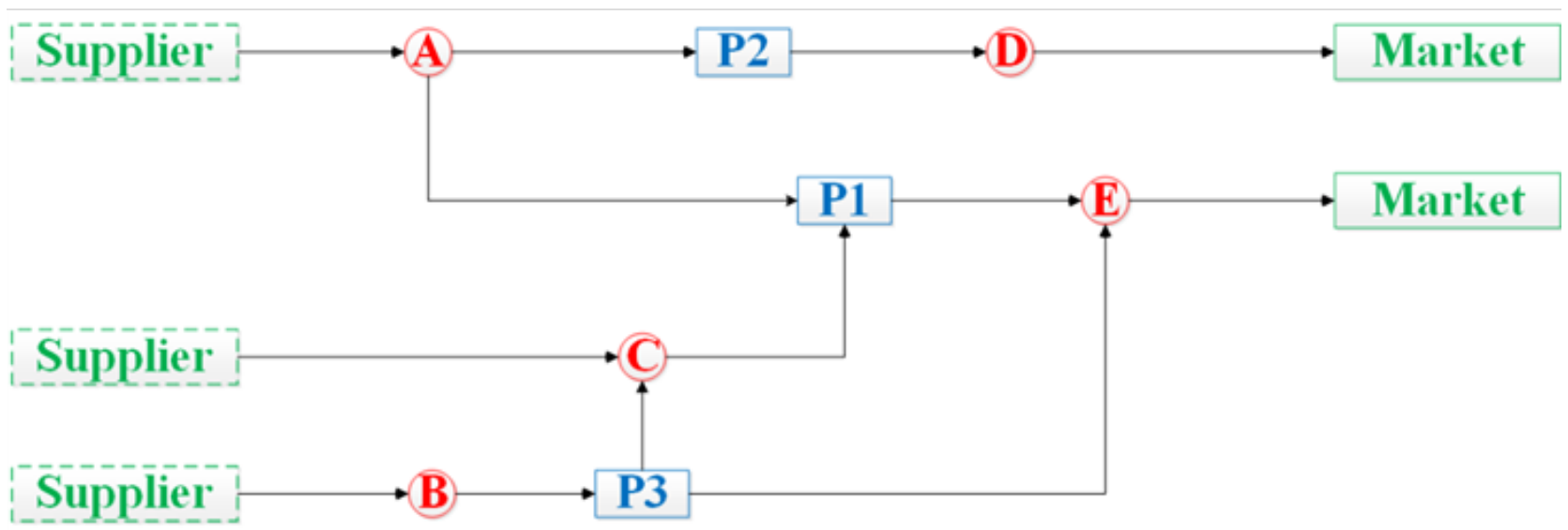

Fig. 5. The chemical process network of case study 1.

The supplies of feedstocks A-C, and demands of products D-E are subject to uncertainty. A total of 160 demand uncertainty data points are used in this case study, and each data point corresponds to a combination of all product demand realizations. For the supply uncertainty, 200 uncertainty data points are used. Each data point represents a combination of supply uncertainty realizations for all raw materials. Demand and supply uncertainty data are attached with the labels of different government policies [67]. Government could encourage or discourage a certain industry by the means of subsidies and tax rates [68]. There are two labels within demand uncertainty data, which indicates two data classes. There are also two data classes within the supply uncertainty data. The two different labels represent two kinds of government policies that encourage or discourage the industry related to the materials and products of this process network [67, 68].

The results of the proposed data-driven uncertainty modeling framework are given in Fig.6 and Fig. 7. In these two figures, the red dots and red stars represent uncertainty data from different data classes, and the polytopes represent data-driven uncertainty sets with uncertainty budgets $\Phi^{dem}$=1 and $\Phi^{sup}$=1. Due to the settings of uncertainty budgets, some data points are not included in the uncertainty sets. The occurrence probabilities of different data classes are inferred from the label information, and the corresponding numerical values are listed in these two figures. From Fig. 6, we can see that the probabilities of demand data classes 1 and 2 are both 0.5. There is only one component in the Dirichlet process mixture model for both data classes. The proposed uncertainty model accurately captures the correlations among the two uncertain parameters for demands. Fig. 7 shows that the occurrence probabilities for the two supply data classes are the same. There is one component in each of these data classes. The DDSRO problem for the planning of process network can be solved using the proposed computational algorithm.

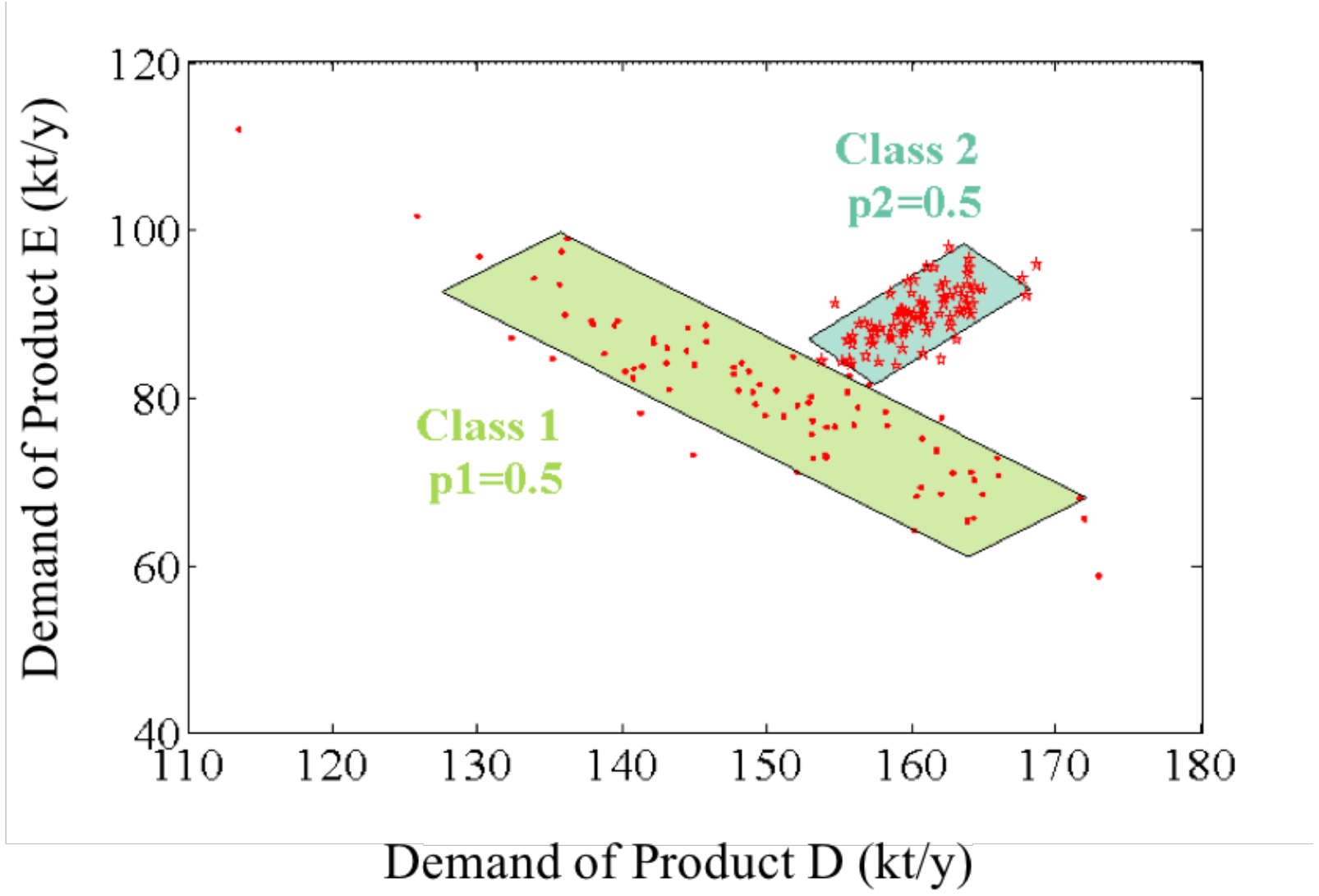

Fig. 6. The data-driven demand uncertainty modeling results ($\Phi^{dem}$=1) for the two data classes in case study 1.

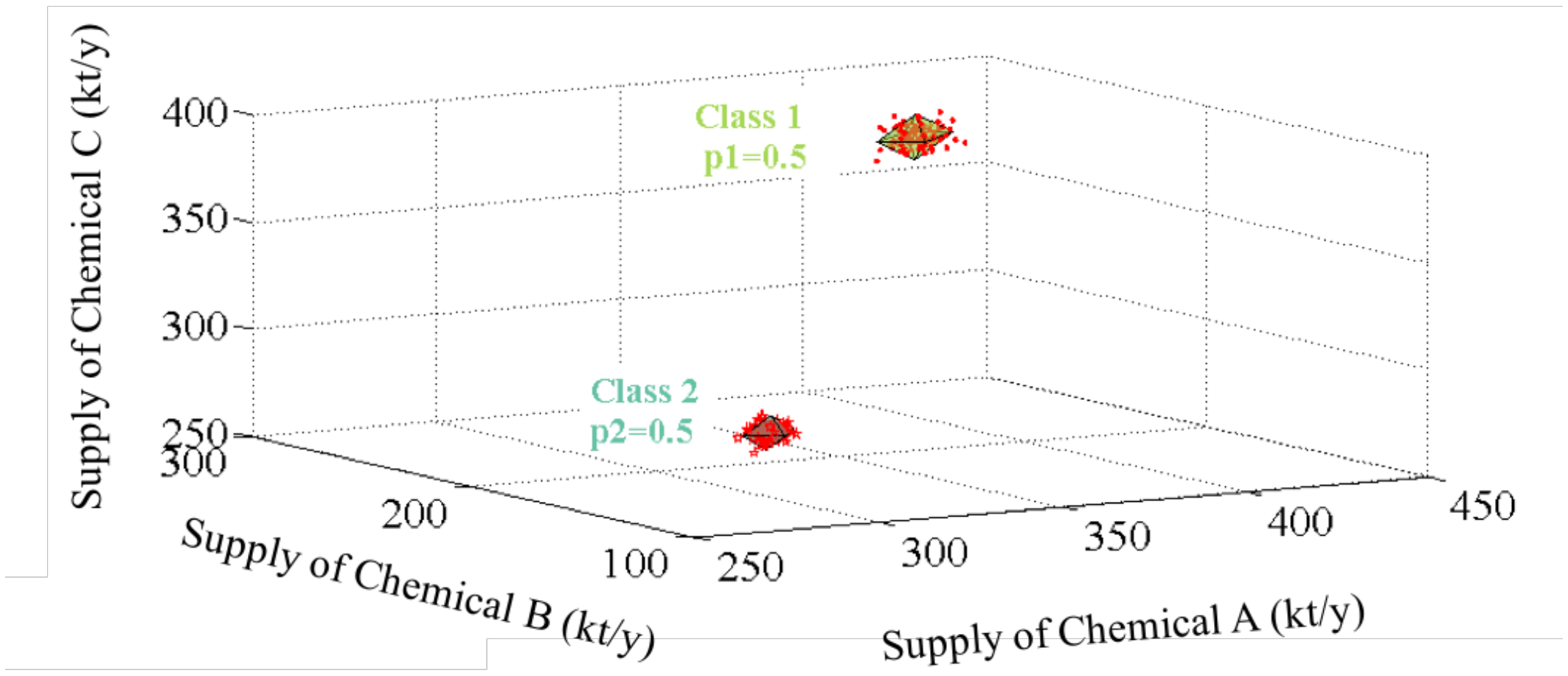

Fig. 7. The data-driven supply uncertainty modeling results ($\Phi^{sup}$=1) for the two data classes in case study 1.

We solve the optimization problem of process network planning using the deterministic optimization method, the two-stage stochastic programming method, the two-stage ARO method with a box set, the DDANRO approach, and the DDSRO approach. All optimization problems are modeled in GAMS 24.7.3 [69], solved with CPLEX 12.6.3, and are implemented on a computer with an Intel (R) Core (TM) i7-6700 CPU @ 3.40 GHz and 32 GB RAM. The relative optimality gaps for both the decomposition-based algorithm in Section 2.4 and the multi-cut Benders decomposition algorithm are set to be 0.1%.

The problem sizes and computational results are summarized in Table 2. For the two-stage stochastic programming problem, each scenario corresponds to a demand uncertainty data point and a supply uncertainty data point. There are 32,000 (160×200) scenarios in the two-stage stochastic programming problem, because there are 160 and 200 data points for demand and supply uncertainty, respectively. Two-stage stochastic programming approach aims to find a solution that performs well on average, so its objective value is only 3.7% higher than that of the DDSRO approach. However, it takes 28 iterations and 106,140 seconds to solve the resulting two-stage stochastic programming problem with the multi-cut Benders decomposition [61]. We note that the deterministic equivalent MILP of this two-stage stochastic program is not suitable to be solved directly due to its large-scale problem size. By contrast, the DDSRO problem can be solved within only 3 seconds, which demonstrates its computational advantage. The two-stage ARO problem with box uncertainty sets is solved using the bounds of demand and supply derived from uncertainty data points in Fig. 6 and Fig. 7, respectively. Data-driven uncertainty sets for demand and supply are constructed based on all the data points in Fig. 6 and Fig. 7, respectively. The uncertainty budgets for demand and supply are both 1. The computational times of the two-stage ARO problem and the DDANRO problem are 1 CPU second and 7 CPU seconds, respectively. In spite of its computational tractability, the two-stage ARO method is conservative and only generates a NPV of $1,372.5MM. Compared with the two-stage ARO method, the DDANRO method ameliorates the conservatism issue and generates $244.6MM higher NPV.

Table 2. Comparisons of problem sizes and computational results in case study 1.

| | Deterministic Planning | Two-stage stochastic programming | Two-stage ARO | DDANRO ($\Delta^{dem}$=1, $\Delta^{sup}$=1)* | DDSRO ($\Phi^{dem}$=1, $\Phi^{sup}$=1)# |
|---|---|---|---|---|---|
| Bin. Var. | 30 | 30 | 30 | 30 | 30 |
| Cont. Var. | 196 | 4,160,060 | 246 | 246 | 468 |
| Constraints | 276 | 10,560,193 | 376 | 396 | 551 |
| Max. NPV ($MM) | 1,809.5 | 1,802.6 | 1,372.5 | 1,671.7 | 1,739.1 |
| Total CPU (s) | 0.2 | 106,140 | 1 | 7 | 3 |
| Iterations | N/A | 28 | 2 | 4 | 2 |

* $\Delta^{dem}$ and $\Delta^{sup}$ are data-driven uncertainty budgets in the DDANRO approach for demand and supply, respectively.

# $\Phi^{dem}$ and $\Phi^{sup}$ are data-driven uncertainty budgets in the DDSRO approach for demand and supply, respectively.

The optimal capacity expansion decisions determined by the two-stage stochastic programming approach, the DDANRO method and DDSRO approach are shown in Fig. 8 (a), (b) and (c), respectively. As can be observed from Fig. 8, Process 1 is expanded at time period 5, and Process 2 starts the expansion at time period 2 for all the three methods. Moreover, Process 2 is further expanded at the 6-th time period in the solution determined by the DDSRO approach. By contrast, Process 2 is not selected to be expanded from time period 2 in the DDANRO solution. The optimal process capacities at the end of the planning horizon determined by the two-stage stochastic programming approach, the DDANRO method and the DDSRO approach are listed in Table 3.

Table 3. The optimal capacity of processes at the end of the planning horizon.

| Process capacity (kt/y) | Two-stage stochastic programming | DDANRO | DDSRO |
|---|---|---|---|
| Process 1 | 100 | 114 | 114 |
| Process 2 | 227 | 207 | 225 |
| Process 3 | 51 | 39 | 39 |

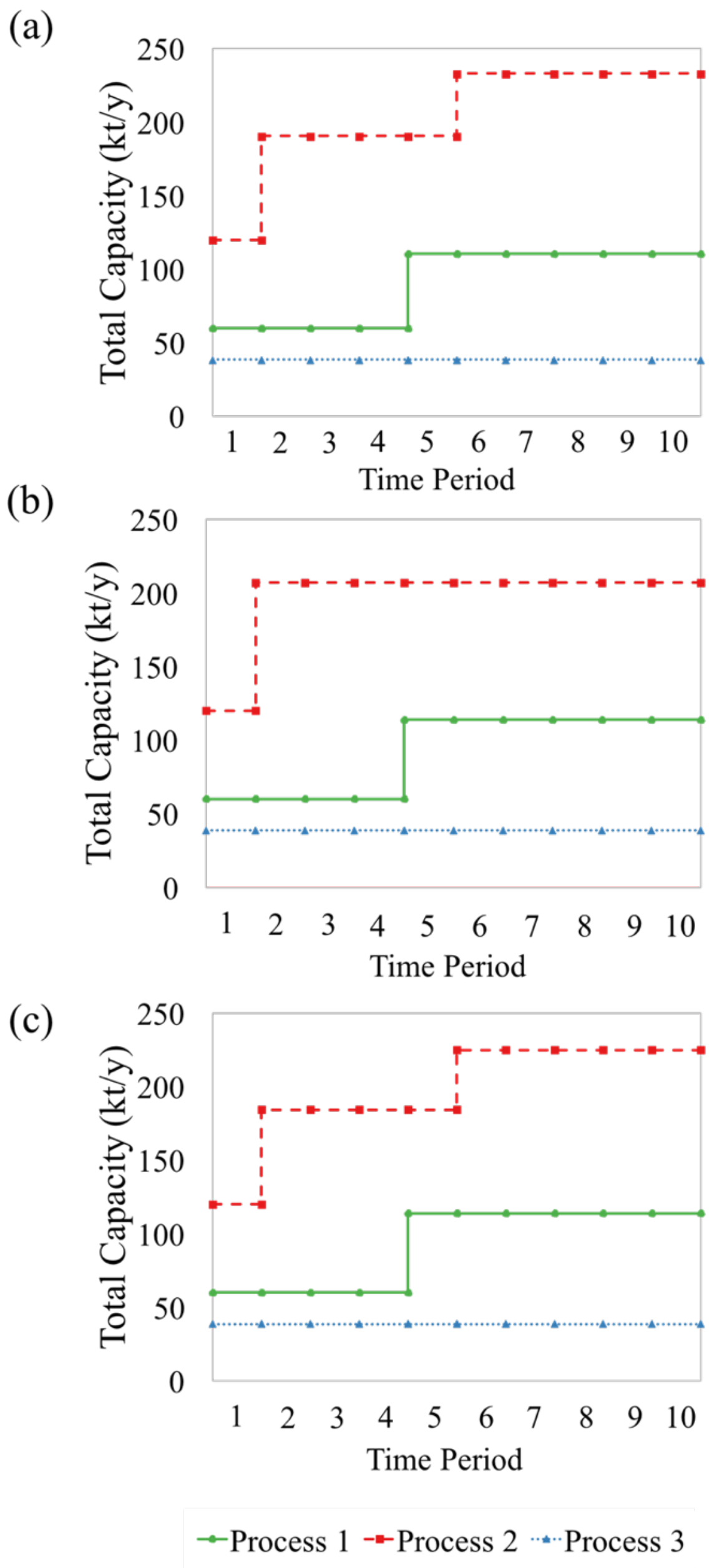

Fig. 8. Optimal capacity expansion decisions over the entire planning horizon determined by (a) the two-stage stochastic programming approach, (b) the DDANRO approach, and (c) the DDSRO approach in case study 1.

## 4.3. Case study 2

A large-scale case study is presented in this subsection to demonstrate the advantages of the DDSRO approach. The process network in this case study consists of 28 chemicals and 38 processes [63], as shown in Fig. 9. Chemicals A-J represent raw materials, which can be purchased from suppliers or manufactured by some processes. Chemicals K-Z are final products, which could be sold to markets. There are 2 intermediates AA and AB. This complex process network has such flexibility that many process technologies are available. For example, Chemical L can be manufactured by Process 2, Process 15 and Process 17. In this case study, we consider 10 time periods over the planning horizon, and the duration of each time period is 2 years. It is assumed that processes 12, 13, 16 and 38 have initial capacities of 40, 25, 300 and 200 kt/y at the beginning of the planning horizon. These 4 processes cannot be expanded until time period 2. The other processes can be built at the beginning of time period 1.

As in the previous case study, supplies of all raw materials and demands of all final products are subject to uncertainty. For the supply uncertainty, a set of 40,000 uncertainty data are used for uncertainty modeling. Each data point has 10 dimensions for a combination of all raw materials. For the demand uncertainty, 40,000 uncertainty data points are used, and each data point is for a combination of all the 16 products. There are 3 labels within demand uncertainty data, which indicates different data classes. There are also 3 data classes within the supply uncertainty data. The 3 labels are different kinds of government policies regarding the industry associated with the process network [67, 68]: (1) encouragement policy, (2) neutral policy and (3) discouragement policy. These policies not only influence the willingness of suppliers to provide related raw materials but also the consumptions of products via subsidies or tax rates. The occurrence probabilities of different data classes are inferred from the label information. The data-driven uncertainty modeling results show that the probabilities of supply data classes 1-3 are 0.25, 0.50, and 0.25, respectively. The probabilities of demand data classes 1-3 are calculated to be 0.25, 0.50, and 0.25, respectively. The number of components in the Dirichlet process mixture models is 1 for all demand data classes. The extracted uncertainty information is incorporated into the DDSRO based process network planning problem.

To demonstrate the advantages of the proposed approach, we solve the process network planning problem using deterministic method, the two-stage stochastic programming approach, the two-stage ARO method, the DDANRO method and the DDSRO approach. In the two-stage stochastic programming problem, there are 1.6 billion scenarios (40,000×40,000), as there are 40,000 data points for both demand and supply uncertainty. In the deterministic equivalent of the two-stage stochastic programming problem, there are 380 integer variables, 1,472,000,001,000 continuous variables, and 1,024,000,004,000 constraints. The size of this two-stage stochastic programming problem is too large to be solved within the computational time limit of 48 hours using the multi-cut Benders decomposition algorithm. While multi-cut Benders decomposition algorithm can take advantage of the problem structure and improve the computational efficiency, it is still intractable to handle this process network planning problem with “big” data for uncertain parameters. Therefore, we only report the problem sizes and computational results of the deterministic approach, the two-stage ARO method, the DDANRO method and the DDSRO approach in Table 4.

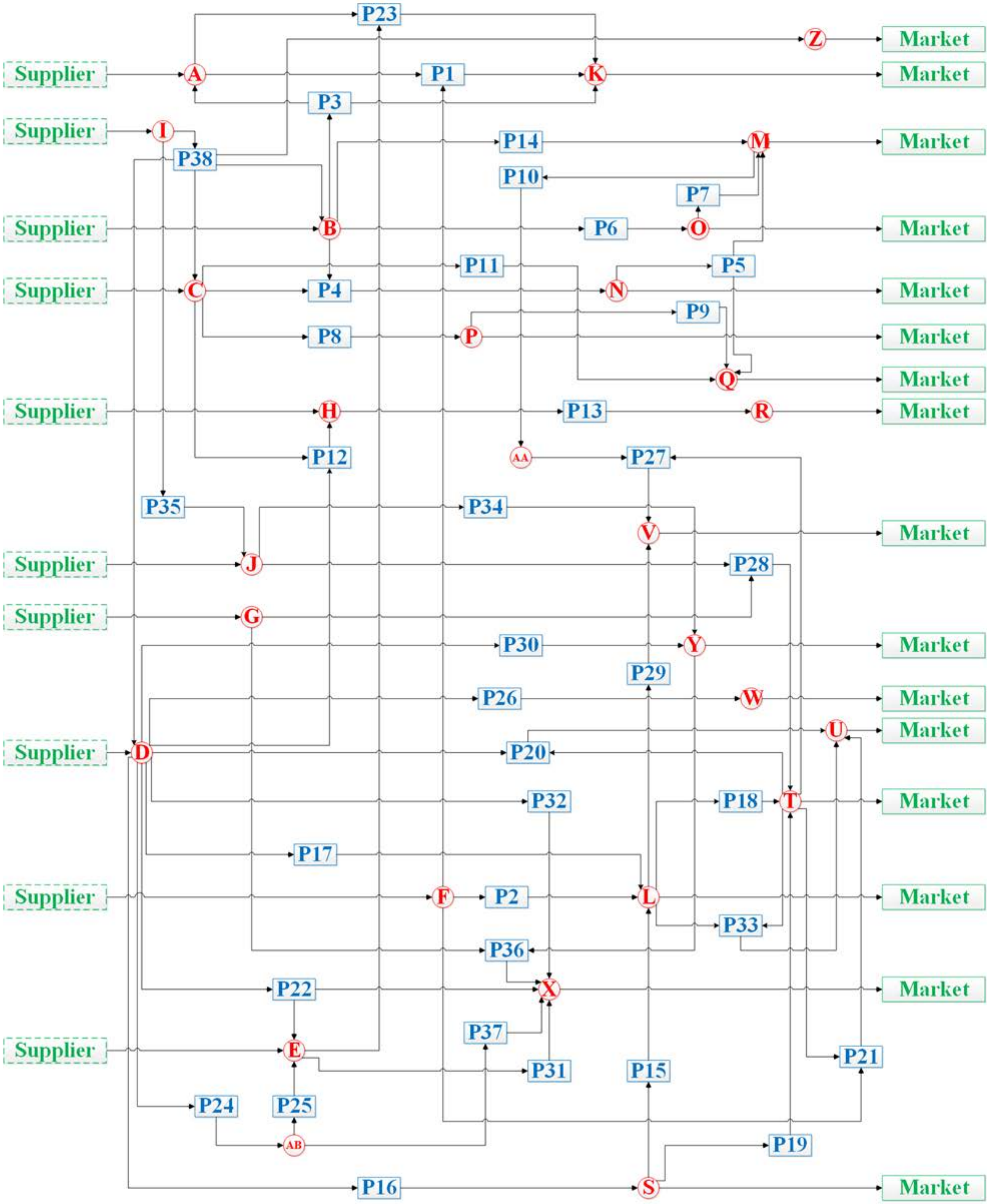

Fig. 9. The chemical process network of case study 2.

Table 4. Comparisons of problem sizes and computational results in case study 2.

| | Deterministic Planning | Two-stage ARO | DDANRO ($\Delta^{dem}$=5, $\Delta^{sup}$=3)* | DDSRO ($\Phi^{dem}$=5, $\Phi^{sup}$=3)# |
|---|---|---|---|---|
| Bin. Var. | 380 | 380 | 380 | 380 |
| Cont. Var. | 1,706 | 1,986 | 1,986 | 3,678 |
| Constraints | 2,366 | 2,926 | 2,946 | 6,462 |
| Max. NPV ($MM) | 2,204.5 | 698.4 | 761.0 | 1,831.2 |
| Total CPU (s) | 0.4 | 5 | 3,232 | 505 |
| Iterations | N/A | 6 | 12 | 6 |

* $\Delta^{dem}$ and $\Delta^{sup}$ are data-driven uncertainty budgets in the DDANRO approach for demand and supply, respective.

\# $\Phi^{dem}$ and $\Phi^{sup}$ are data-driven uncertainty budgets in the DDSRO approach for demand and supply, respective.

From Table 4, we can see that the deterministic planning method consumes less computational time, and generates the highest NPV. However, its solution suffers from infeasibility issue if the supply of raw materials and demand of final products are uncertain. By leveraging the corresponding strengths of two-stage stochastic programming approach and the two-stage ARO method, the DDSRO approach generates $1,070.2MM higher NPV than the solution determined by the DDANRO method, while still maintaining computational tractability. The DDSRO problem can be solved using only 505 seconds. The optimal design and planning decisions at the end of the planning horizon determined by the deterministic optimization method, DDANRO, and DDSRO are shown in Fig. 10, Fig. 11 and Fig. 12, respectively. In these 3 figures, the optimal total production capacities are displayed under operating processes.

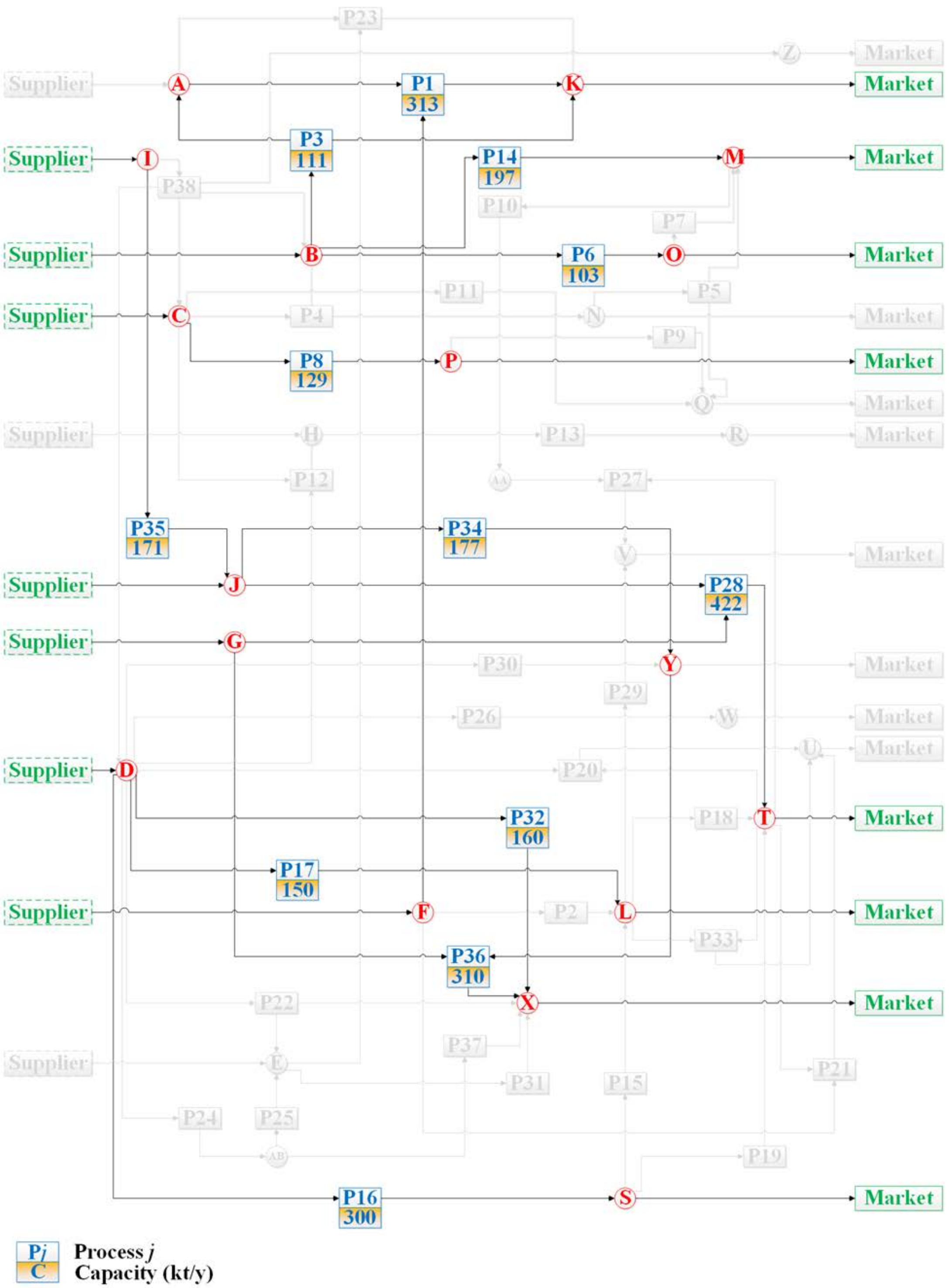

Fig. 10. The optimal design and planning decisions at the end of the planning horizon determined by the deterministic optimization method in case study 2.

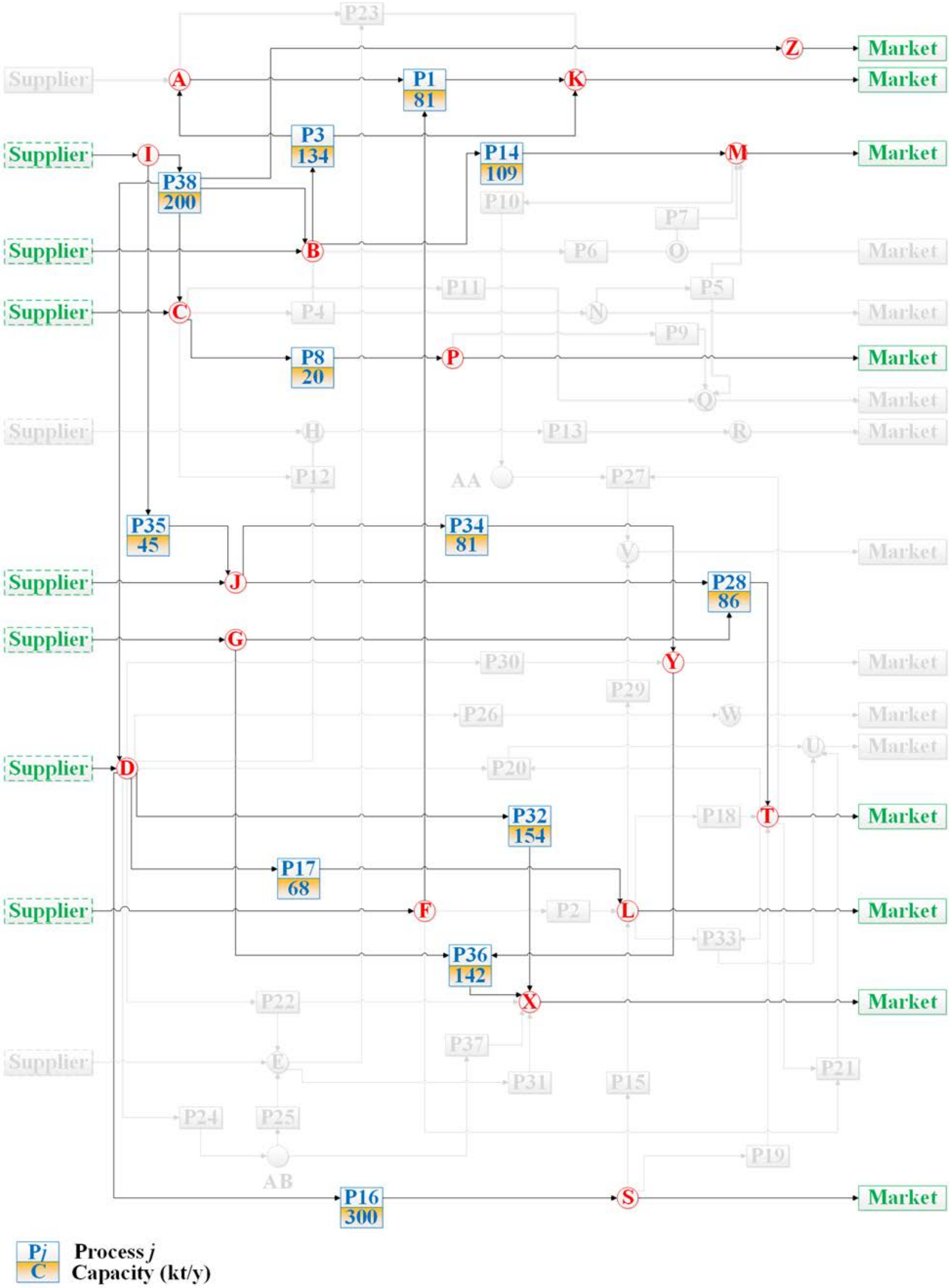

Fig. 11. The optimal design and planning decisions at the end of the planning horizon determined by the DDANRO approach in case study 2.

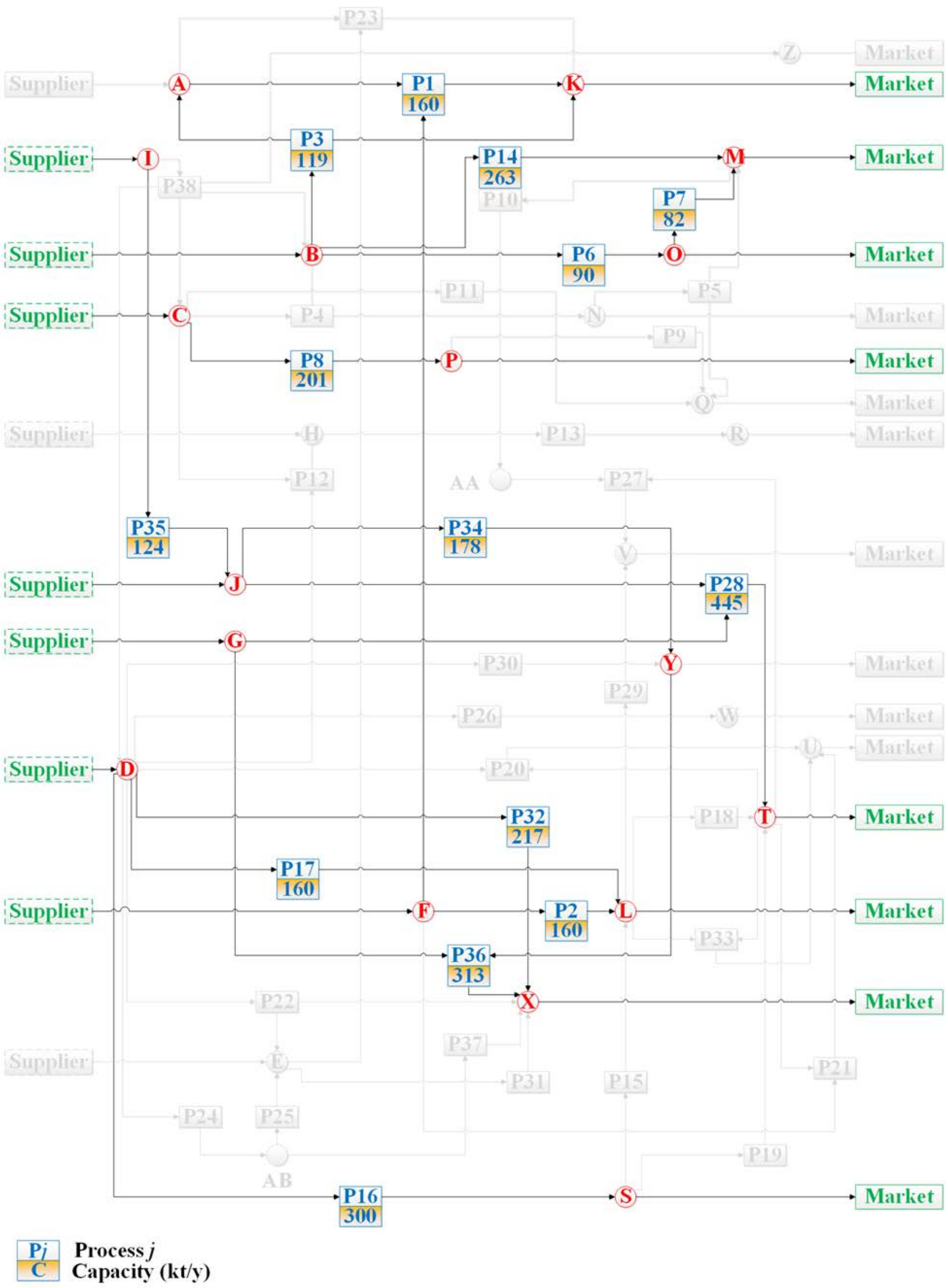

Fig. 12. The optimal design and planning decisions at the end of the planning horizon determined by the proposed DDSRO approach in case study 2.

The optimal capacity expansion decisions determined by the DDANRO method and the DDSRO approach are shown in Fig. 13 and Fig. 14, respectively. From Fig. 13, we can see that a total of 15 processes are selected in the optimal process network determined by the DDANRO method. As shown in Fig. 14, a total of 18 processes are chosen in the optimal process network determined by the DDSRO approach. Specifically, Processes 2, 6 and 11 are not selected in the optimal process network determined by the DDANRO method. Besides, the optimal expansion frequencies and total capacities of Process 28 determined by the two methods are different.

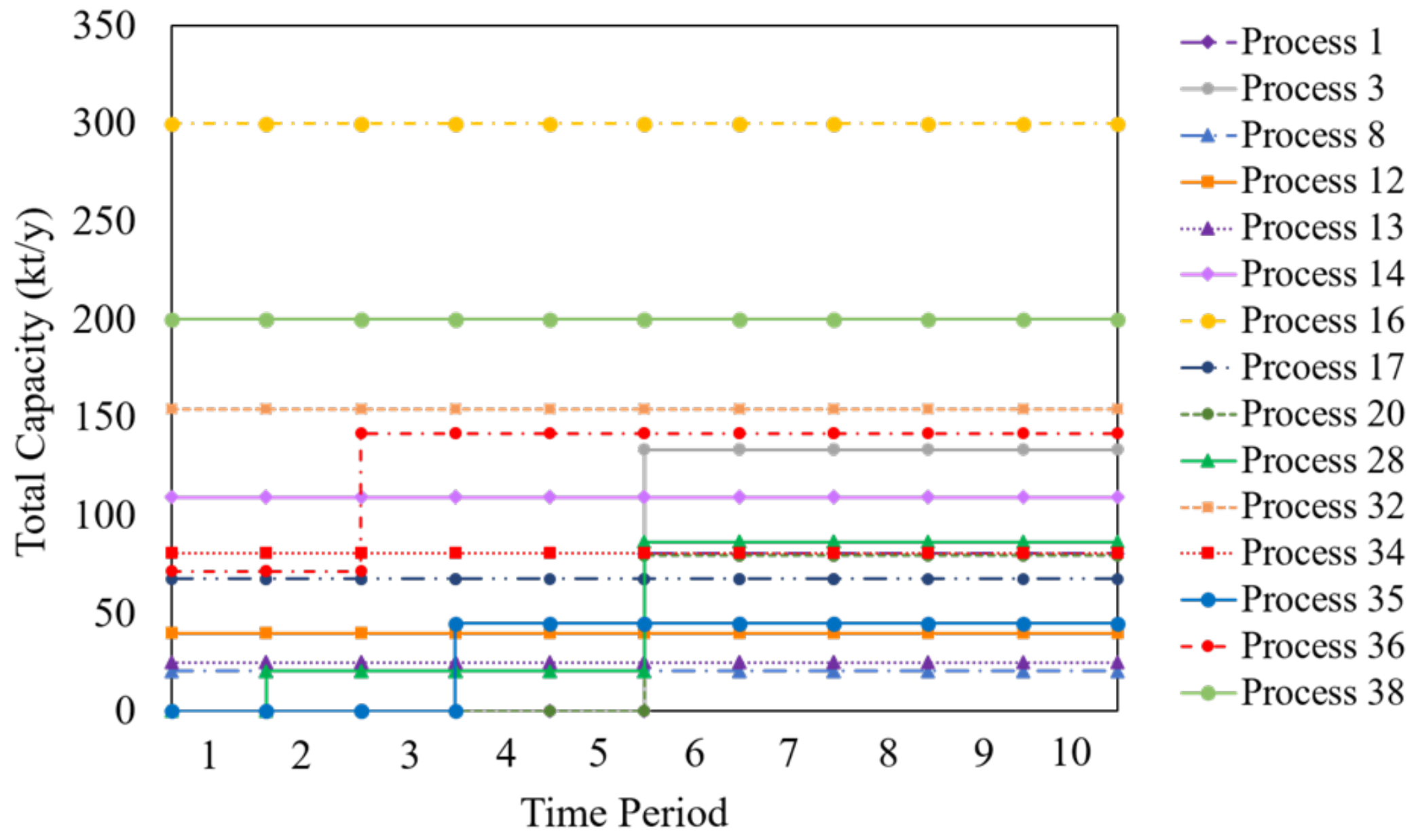

Fig. 13. Optimal capacity expansion decisions over the entire planning horizon determined by the DDANRO approach in case study 2.

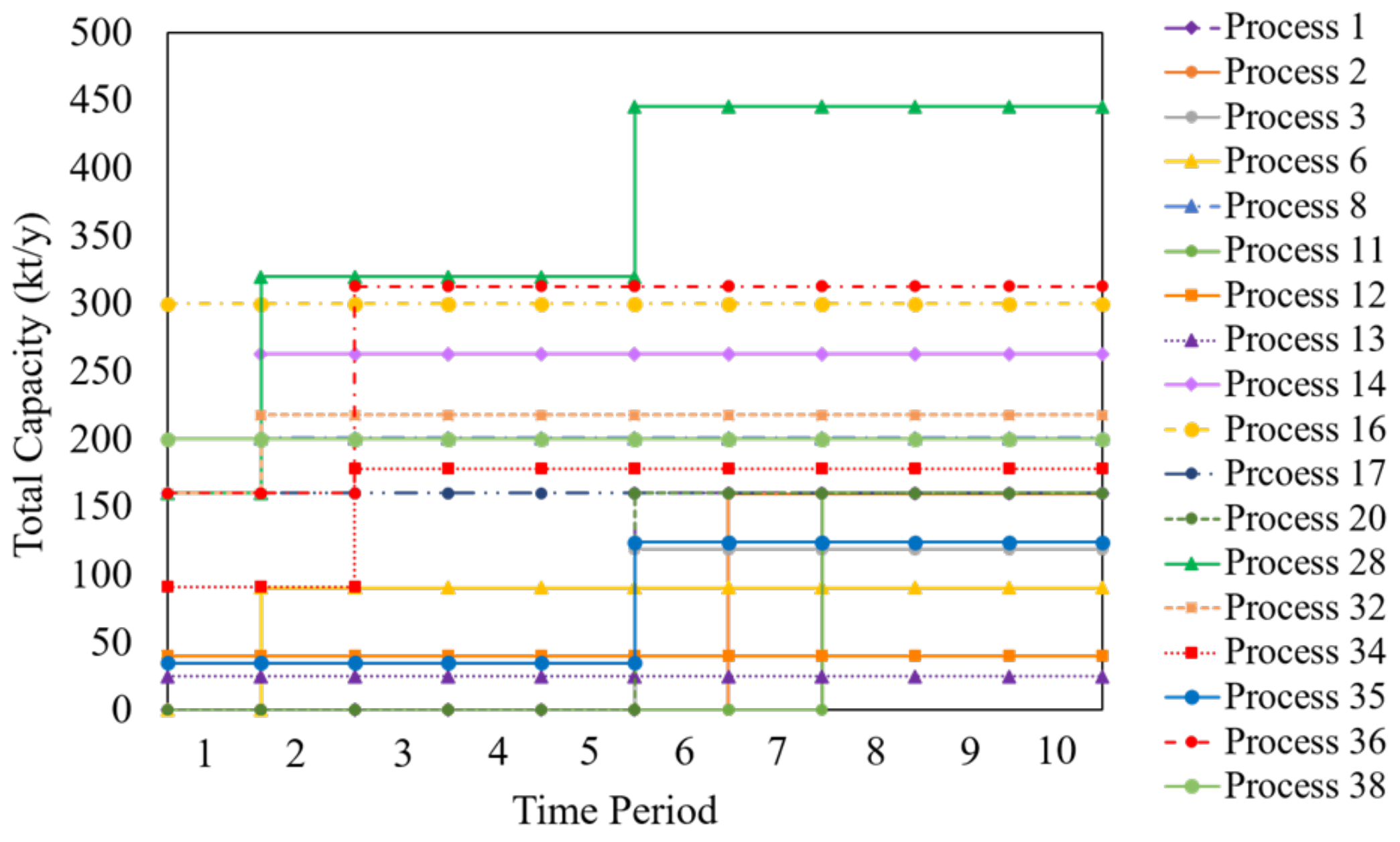

Fig. 14. Optimal capacity expansion decisions over the entire planning horizon determined by the proposed DDSRO approach in case study 2.

Fig. 15 displays the upper and lower bounds in each iteration of the proposed algorithm. The X-axis and Y-axis denote the iteration number and the objective function values, respectively. In Fig. 15, the green dots stand for the upper bounds, while the yellow circles represent the lower bounds. The proposed solution algorithm requires only 6 iterations to converge. During the first 3 iterations, the relative optimality gap decreases significantly from 88.1% to 0.6%.

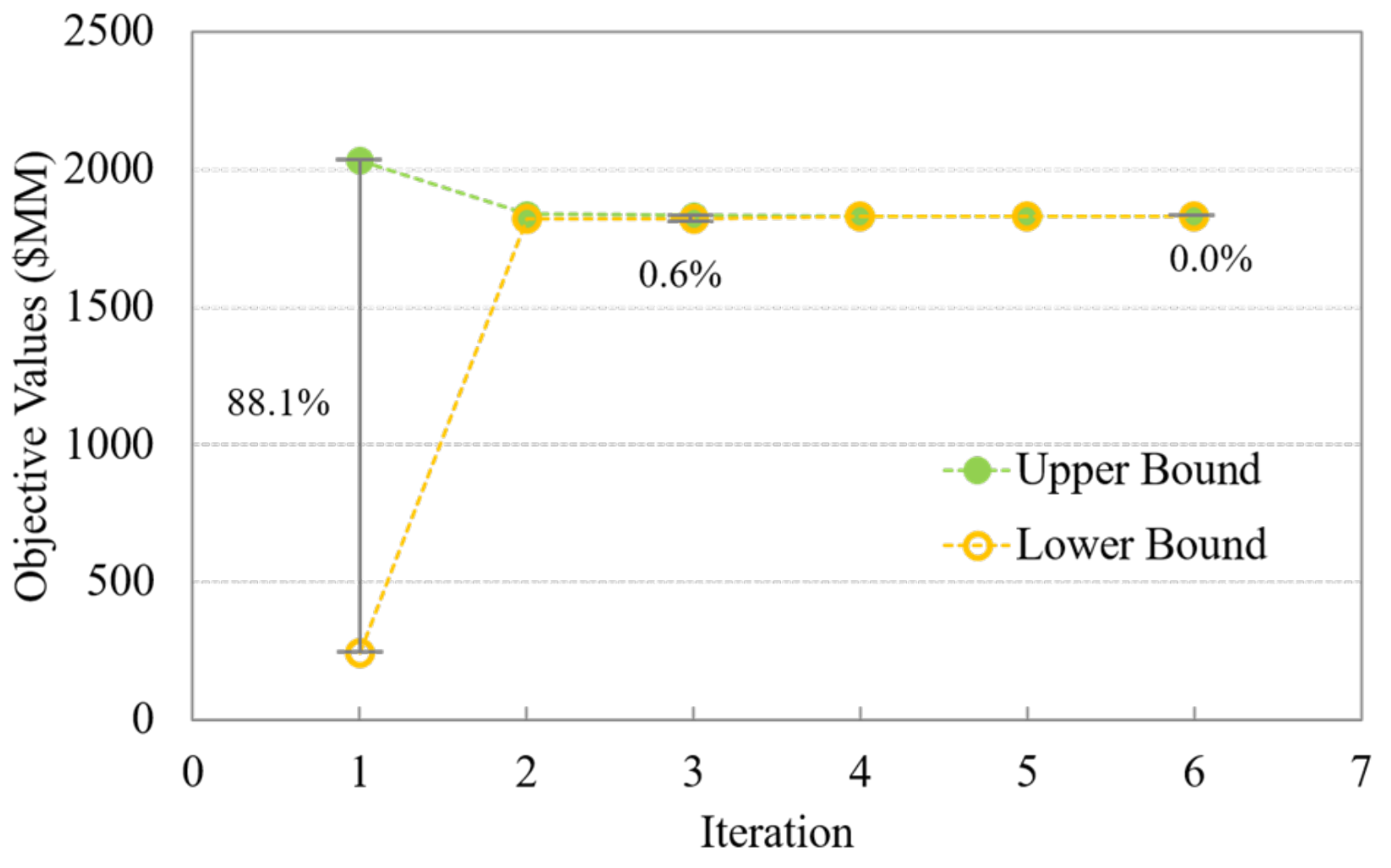

Fig. 15. Upper and lower bounds in each iteration of the decomposition-based algorithm in case study 2.

## 5. Conclusions

This paper proposed a novel DDSRO framework that leveraged machine learning methods to extract accurate uncertainty information from multi-class uncertainty data for decision making under uncertainty. The probability distribution of data classes was learned through maximum likelihood estimation. A group of Dirichlet process mixture models were used to construct uncertainty sets for each data class. Stochastic programming approach was nested as the outer optimization problem to leverage available probability distribution, while ARO was nested as the inner optimization problem for computational tractability. The proposed DDSRO framework leveraged the strengths of both stochastic programming and robust optimization. A decomposition-based algorithm was further developed to efficiently solve the resulting DDSRO problem. Two-stage stochastic programming approach was computationally intractable for the large-scale optimization problem with huge amounts of uncertainty data. The DDSRO approach was much more computationally efficient than two-stage stochastic programming approach. The size of the stochastic programming problem grew exponentially as the number of uncertainty data points increased. By contrast, the problem size of DDSRO was less sensitive to the amount of uncertainty data. Therefore, DDSRO remained tractable to handle problems with "big" data for uncertainties. Moreover, DDSRO generated less conservative solutions compared with two-stage ARO method. Moreover, the DDSRO approach could be considered as a generalization of the DDANRO method. The results showed that the DDSRO approach was advantageous over the DDANRO method for labeled multi-class uncertainty data.

## Acknowledgements

The authors acknowledge financial support from the National Science Foundation (NSF) CAREER Award (CBET-1643244).

## Appendix A. Dirichlet Process Mixture Model

The Dirichlet process is a widely used stochastic process, especially in the Dirichlet process mixture model. Suppose a random distribution $G$ follows a Dirichlet process, denoted as $G \sim DP$ (α, $G_0$), ($G(\Theta_1)$,…, $G(\Theta_r)$) is then Dirichlet distributed for any measurable partition ($\Theta_1$, ..., $\Theta_r$) of Θ [70].

$$\left(G(\Theta_1),\ldots,G(\Theta_r)\right) \sim Dir\left(\alpha G_0(\Theta_1),\ldots,\alpha G_0(\Theta_r)\right) \tag{A1}$$

where α and $G_0$ are the concentration parameter and base distribution, respectively. *Dir* in (A1) represents the Dirichlet distribution. A random draw from the Dirichlet process follows the stick-breaking construction in (A2) [71].

$$G = \sum\nolimits_{k=1}^{\infty} \pi_k \delta_{\theta_k} \tag{A2}$$

where $\pi_k = \beta_k \prod\nolimits_{j=1}^{k-1}\left(1-\beta_j\right)$, $\beta_k$ is the proportion of stick broken off from the remaining stick, and $\beta_k \sim Beta\,(1, \alpha)$. $\theta_k$ is the random variable drawn from the base distribution $G_0$. The delta function $\delta_{\theta_k}$ is equal to 1 at the location of $\theta_k$, and becomes 0 elsewhere.

In the Dirichlet process mixture model, $\theta_k$ is treated as the parameters of the data distribution. A significant feature of the Dirichlet process mixture model is that it has a potentially infinite number of components. This novel feature grants it with great flexibility to model intricate data distribution. The Dirichlet process mixture model is given in (A3) [58].

$$\begin{aligned}
&\beta_k \mid \alpha \sim Beta(1,\alpha) \\
&\theta_k \mid G_0 \sim G_0 \\
&t_i \mid \{\beta_1,\beta_2,\ldots\} \sim Mult\left(\pi(\beta)\right) \\
&o_i \mid t_i \sim p(o_i \mid \theta_{t_i})
\end{aligned} \tag{A3}$$

where *Beta* and *Mult* denote Beta distribution and multinomial distribution, respectively. The observation is denoted as $o_i$ drawn from the $t_i$ th component.

## Appendix B. Details of The Decomposition Algorithm

The multi-level optimization problem (10) can be reformulated to an equivalent single-level optimization problem by enumerating all the extreme points [72]. However, this single-level optimization problem could be a large-scale optimization problem due to the potentially large number of extreme points. Therefore, we partially enumerate the extreme points, and construct the master problem following the literature [8]. Because master problem **(MP)** contains a subset of the extreme points, it is a relaxation of the original single-level optimization problem. The master problem **(MP)** is shown below.

$$\textbf{(MP)}\qquad \begin{aligned} \min_{\mathbf{x},\eta,\mathbf{y}_s^l}\ & \mathbf{c}^{\mathrm{T}}\mathbf{x}+\eta \\ \text{s.t. }\ & \mathbf{A}\mathbf{x}\ge \mathbf{d} \\ & \eta \ge \sum_s p_s\left(\mathbf{b}^{\mathrm{T}}\mathbf{y}_s^l\right),\ \ l=1,\ldots,r \\ & \mathbf{T}\mathbf{x}+\mathbf{W}\mathbf{y}_s^l \ge \mathbf{h}-\mathbf{M}\mathbf{u}_s^l,\ \ l=1,\ldots,r,\forall s \\ & \mathbf{x}\in R_+^{n_1}\times Z^{n_2},\ \ \mathbf{y}_s^l\in R_+^{n_3},\ \ l=1,\ldots,r,\forall s \end{aligned}$$

where $s$ is the index for data class, and $\mathbf{u}_s^l$ is the enumerated uncertainty realization at the $l$ th iteration for data class $s$, $\mathbf{y}_s^l$ is its corresponding recourse variable, and $r$ stands for the current number of iterations.

To enumerate the important uncertainty realization on-the-fly, the sub-problem in (B1) is solved in each iteration [8].

$$\begin{aligned} Q_{s,i}(\mathbf{x})=\max_{\mathbf{u}\in U_{s,i}}\ & \min_{\mathbf{y}_s}\ \mathbf{b}^{\mathrm{T}}\mathbf{y}_s \\ & \text{s.t. }\ \mathbf{W}\mathbf{y}_s \ge \mathbf{h}-\mathbf{T}\mathbf{x}-\mathbf{M}\mathbf{u} \\ & \qquad \mathbf{y}_s\in R_+^{n_3} \end{aligned} \tag{B1}$$

It is worth noting that (B1) is a max-min optimization problem, which can be transformed to a single-level optimization problem by employing strong duality or KKT conditions. To make the sub-problem computationally tractable, we dualize the inner optimization problem of (B1), which is a linear program with regard to $\mathbf{y}_s$, and merge the dual problem with the maximization problem with respect to $\mathbf{u}$. The resulting problem is shown below.

$$(\mathbf{SUP}_{s,i})\qquad \begin{aligned} Q_{s,i}(\mathbf{x})=\max_{\mathbf{z},\boldsymbol{\varphi}}\ & \left(\mathbf{h}-\mathbf{T}\mathbf{x}-\mathbf{M}\boldsymbol{\mu}_{s,i}\right)^{\mathrm{T}}\boldsymbol{\varphi}-\sum_t\sum_j\left(\kappa_{s,i}\mathbf{M}\boldsymbol{\Psi}_{s,i}^{1/2}\Lambda_{s,i}\right)_{tj}\mathbf{z}_j\boldsymbol{\varphi}_t \\ \text{s.t. }\ & \mathbf{W}^{\mathrm{T}}\boldsymbol{\varphi}\le \mathbf{b} \\ & \boldsymbol{\varphi}\ge \mathbf{0} \\ & \|\mathbf{z}\|_\infty\le 1,\ \ \|\mathbf{z}\|_1\le \Phi_{s,i} \end{aligned}$$

where $\boldsymbol{\varphi}$ is the vector of dual variables corresponding to the constraint $\mathbf{W}\mathbf{y}\ge\mathbf{h}-\mathbf{T}\mathbf{x}-\mathbf{M}\mathbf{u}$, $\boldsymbol{\varphi}_t$ represents the $t$ th component in vector $\boldsymbol{\varphi}$, and $\mathbf{z}_j$ is the $j$ th component in vector $\mathbf{z}$. Problem $(\mathbf{SUP}_{s,i})$ is reformulated to handle the bilinear term $\mathbf{z}_j\boldsymbol{\varphi}_t$ in its objective function.

To facilitate the reformulation, $\mathbf{z}_j$ is divided into two parts $\mathbf{z}_j=\mathbf{z}_j^+-\mathbf{z}_j^-$. Thus, we have

$$\mathbf{u}=\boldsymbol{\mu}_{s,i}+\kappa_{s,i}\boldsymbol{\Psi}_{s,i}^{1/2}\Lambda_{s,i}\left(\mathbf{z}^+-\mathbf{z}^-\right) \tag{B2}$$

Following the existing literature [73, 74], the uncertainty budget parameter $\Phi_{s,i}$ is typically set as an integer value due to its physical meaning. For the integer uncertainty budget, the optimal $\mathbf{z}_j^+$ and $\mathbf{z}_j^-$ take the values of 0 or 1, and these two variables can be restricted to binary variables. We employ Glover's linearization for the bilinear terms $\mathbf{G}_{tj}^+ = \mathbf{z}_j^+ \boldsymbol{\varphi}_t$ and $\mathbf{G}_{tj}^- = \mathbf{z}_j^- \boldsymbol{\varphi}_t$ [75], which are the products of a binary variable $\mathbf{z}_j^+$ (and $\mathbf{z}_j^-$) and a continuous variable $\boldsymbol{\varphi}_t$, and reformulate **(SUP$_{s,i}$)** into (B3) following the procedure introduced in [8].

$$
\begin{aligned}
Q_{s,i}(\mathbf{x}) \quad = & \max_{\boldsymbol{\varphi},\mathbf{z}^+,\mathbf{z}^-,\mathbf{G}^+,\mathbf{G}^-} \left(\mathbf{h}-\mathbf{Tx}-\mathbf{M}\boldsymbol{\mu}_{s,i}\right)^{\mathrm{T}} \boldsymbol{\varphi} - \mathrm{Tr}\left(\left(\kappa_{s,i}\mathbf{M}\boldsymbol{\Psi}_{s,i}^{1/2}\Lambda_{s,i}\right)^{\mathrm{T}}\left(\mathbf{G}^+ - \mathbf{G}^-\right)\right) \\
& \text{s.t. } \mathbf{W}^T\boldsymbol{\varphi} \le \mathbf{b} \\
& \quad \boldsymbol{\varphi} \ge \mathbf{0} \\
& \quad \mathbf{0} \le \mathbf{G}^+ \le \boldsymbol{\varphi}\cdot\mathbf{e}^{\mathrm{T}} \\
& \quad \mathbf{0} \le \mathbf{G}^- \le \boldsymbol{\varphi}\cdot\mathbf{e}^{\mathrm{T}} \\
& \quad \mathbf{G}^+ \le \tilde{\mathbf{e}}\cdot\left(M_0\cdot\mathbf{z}^+\right)^{\mathrm{T}} \\
& \quad \mathbf{G}^- \le \tilde{\mathbf{e}}\cdot\left(M_0\cdot\mathbf{z}^-\right)^{\mathrm{T}} \\
& \quad \mathbf{G}^+ \ge \boldsymbol{\varphi}\cdot\mathbf{e}^{\mathrm{T}} - \tilde{\mathbf{e}}\cdot\left(M_0\cdot\left(\mathbf{e}-\mathbf{z}^+\right)\right)^{\mathrm{T}} \\
& \quad \mathbf{G}^- \ge \boldsymbol{\varphi}\cdot\mathbf{e}^{\mathrm{T}} - \tilde{\mathbf{e}}\cdot\left(M_0\cdot\left(\mathbf{e}-\mathbf{z}^-\right)\right)^{\mathrm{T}} \\
& \quad \mathbf{e}^{\mathrm{T}}\left(\mathbf{z}^+ + \mathbf{z}^-\right) \le \Phi_{s,i} \\
& \quad \mathbf{z}^+ + \mathbf{z}^- \le \mathbf{e}, \quad \mathbf{z}^+, \mathbf{z}^- \in \{0,1\}^K
\end{aligned}
\tag{B3}
$$

where $\mathbf{e}$ and $\tilde{\mathbf{e}}$ denote vectors with all elements being one. The $t$ th row and $j$ th column elements of matrices $\mathbf{G}^+$ and $\mathbf{G}^-$ are $\mathbf{G}_{tj}^+$ and $\mathbf{G}_{tj}^-$, respectively.

## Nomenclature

### *DDSRO Framework*

The main notations used in the DDSRO modeling framework are listed below.

| | |
|---|---|
| **b** | vector of objective coefficients corresponding to second-stage decisions |
| **c** | vector of objective coefficients corresponding to first-stage decisions |
| **u** | vector of uncertainties |
| **x** | vector of first-stage decisions |
| $\mathbf{y}_s$ | vector of second-stage decisions for data class *s* |
| $C$ | total number of data classes in uncertainty data |
| $m(s)$ | total number of basic uncertainty sets for data class $s$ |
| $M$ | truncation level in the Variational inference |
| $M_0$ | constant with a very large value |
| $U_s$ | uncertainty set for data class $s$ |
| $s$ | index of data class |
| $\Lambda_i$ | scaling factor in the data-driven uncertainty set |
| $\tau_i$ | a hyper-parameter for Beta distribution in the Variational inference |
| $\nu_i$ | a hyper-parameter for Beta distribution in the Variational inference |
| $\gamma_i$ | weight of mixture component |
| $\gamma^*$ | threshold value for the weight of mixture component |
| $\Phi_{s,i}$ | uncertainty budget in data-driven uncertainty set |
| $\mathbf{A}$ | coefficient matrix corresponding to the first-stage decisions |
| $\mathbf{M}$ | coefficient matrix corresponding to the uncertainties |
| $\mathbf{T}$ | technology matrix in the DDSRO framework |
| $\mathbf{W}$ | recourse matrix in the DDSRO framework |

***Application to planning of process networks***

The sets, parameters, and variables used in the application part are summarized below. Note that all parameters are denoted in lower-case symbols, and all variables are denoted in upper-case symbols.

*Sets/indices*

| | |
|---|---|
| $I$ | set of processes indexed by $i$ |
| $J$ | set of chemicals indexed by $j$ |
| $T$ | set of time periods indexed by $t$ |
| $\Xi$ | set of data classes indexed by $s$ |

*Parameters*

| | |
|---|---|
| $c1_{it}$ | variable investment cost for process $i$ in time period $t$ |

$c2_{it}$ fixed investment cost for process $i$ in time period $t$

$c3_{it}$ unit operating cost for process $i$ in time period $t$

$c4_{it}$ purchase price of chemical $j$ in time period $t$

$cb_t$ maximum allowable investment in time period $t$

$ce_i$ maximum number of expansions for process $i$ over the planning horizon

$du_{jt}$ demand of chemical $j$ in time period $t$

$qe_{it}^L$ lower bound for capacity expansion of process $i$ in time period $t$

$qe_{it}^U$ upper bound for capacity expansion of process $i$ in time period $t$

$su_{jt}$ supply of chemical $j$ in time period $t$

$v_{jt}$ sale price of chemical $j$ in time period $t$

$\kappa_{ij}$ mass balance coefficient for chemical $j$ in process $i$

*Binary variables*

$Y_{it}$ binary variable that indicates whether process $i$ is chosen for expansion in time period $t$

*Continuous variables*

$P_{sjt}$ purchase amount of chemical $j$ in time period $t$ for data class $s$

$Q_{it}$ total capacity of process $i$ in time period $t$

$QE_{it}$ capacity expansion of process $i$ in time period $t$

$SA_{sjt}$ sale amount of chemical $j$ in time period $t$ for data class $s$

$W_{sit}$ operation level of process $i$ in time period $t$ for data class $s$